\documentclass[journal]{IEEEtran}
\usepackage{ifpdf}
\usepackage{cite}
\ifCLASSINFOpdf
  \usepackage[pdftex]{graphicx}
  \usepackage{subfigure}
\else
\fi
\usepackage{amsmath}
\usepackage{multirow}
\usepackage{pifont}
\usepackage[colorlinks]{hyperref}
\usepackage{algorithmic}
\begin{document}
\title{Progressive Unsupervised Person Re-identification by Tracklet Association with Spatio-Temporal Regularization}

\author{Qiaokang~Xie, 
        Wengang~Zhou, 
        Guo-Jun~Qi,~\IEEEmembership{Member,~IEEE,}
        Qi~Tian,~\IEEEmembership{Fellow,~IEEE,}
        and~Houqiang~Li,~\IEEEmembership{Senior~Member,~IEEE}% <-this % stops a space
%\author{Qiaokang~Xie,~\IEEEmembership{Member,~IEEE,}
%	John~Doe,~\emph{i.e.}EEmembership{Fellow,~OSA,}
%	and~Jane~Doe,~\emph{i.e.}EEmembership{Life~Fellow,~IEEE}% <-this % stops a space
\thanks{Qiaokang~Xie, Wengang~Zhou, and Houqiang~Li are with the CAS Key Laboratory of Technology in Geo-spatial Information Processing and Application System, Department of Electronic Engineering and Information Science, University of Science and Technology of China, Hefei, 230027, China. E-mail: xieqiaok@mail.ustc.edu.cn, \{zhwg, lihq\}@ustc.edu.cn.}% <-this % stops a space
\thanks{Guo-Jun~Qi is with Huawei Cloud EI Product Department.\quad E-mail: guojun.qi@huawei.com.}% <-this % stops a space
\thanks{Qi~Tian is with Huawei Noah's Ark Laboratory. E-mail: tian.qi1@huawei.com.}% <-this % stops a space
\thanks{Corresponding authors: Wengang~Zhou and Houqiang~Li.}
}

% note the % following the last \emph{i.e.}EEmembership and also \thanks - 
% Spaces after \emph{i.e.}EEmembership other than the last one are OK (and needed) as

\markboth{IEEE Transactions on Multimedia,~Vol.~**, No.~**, August~20**}%
{Shell \MakeLowercase{\textit{et al.}}: Bare Demo of IEEEtran.cls for IEEE Journals}

%\emph{i.e.}EEpubid{0000--0000/00\$00.00~\copyright~2015 IEEE}
% Remember, if you use this you must call \emph{i.e.}EEpubidadjcol in the second


\maketitle

\begin{abstract}
  Existing methods for person re-identification (Re-ID) are mostly based on supervised learning which requires numerous manually labeled samples across all camera views for training. Such a paradigm suffers the scalability issue since in real-world Re-ID application, it is difficult to exhaustively label abundant identities over multiple disjoint camera views. To this end, we propose a progressive deep learning method for unsupervised person Re-ID in the wild by Tracklet Association with Spatio-Temporal Regularization (TASTR). In our approach, we first collect tracklet data within each camera by automatic person detection and tracking. Then, an initial Re-ID model is trained based on within-camera triplet construction for person representation learning. After that, based on the person visual feature and spatio-temporal constraint, we associate cross-camera tracklets to generate cross-camera triplets and update the Re-ID model. Lastly, with the refined Re-ID model, better visual feature of person can be extracted, which further promote the association of cross-camera tracklets. The last two steps are iterated multiple times to progressively upgrade the Re-ID model. 
%  It automatically mines the identity discriminative information from person tracklet data and spatio-temporal topology of the camera network. 
  To facilitate the study, we have collected a new 4K UHD video dataset named Campus-4K with full frames and full spatio-temporal information. Experimental results show that with the spatio-temporal constraint in the training phase, the proposed approach outperforms the state-of-the-art unsupervised methods by notable margins on DukeMTMC-reID, and achieves competitive performance to fully supervised methods on both DukeMTMC-reID and Campus-4K datasets. 
\end{abstract}


% \fi
\IEEEpeerreviewmaketitle

\section{Introduction}
% \emph{i.e.}EEPARstart{A}{demo} file is ....
% \emph{i.e.}EEPARstart{A}{}demo file is ....
% \emph{i.e.}EEPARstart{T}{his demo} file is ....
\IEEEPARstart{A}{s} a hot topic in computer vision, person re-identification (Re-ID) aims at matching pedestrians detected from non-overlapping camera views. Thanks to the potential significance in video surveillance applications, it has attracted wide attention from both academia and industry. In person Re-ID, the key is to learn a good feature representation for pedestrian, which is expected to be invariant to view, pose and the change of cameras, \emph{etc.} In recent years, the performance of person Re-ID has been greatly boosted with the development of deep learning techniques and the release of many large-scale public datasets. 

Most existing approaches~\cite{li2014deepreid,wang2015zero,ahmed2015improved,ye2016person,chen2017beyond,Zhao_2017_ICCV,hermans2017defense,liu2017end,zhou2017large,li2018harmonious,wei2018person,song2018mask,wang2019incremental,zheng2019pose} for person Re-ID follow the supervised learning paradigm on labeled datasets, where cross-view identity matching image pairs are supposed to be manually labeled for each camera pair. However, we may suffer performance degradation when directly deploying these trained models to a different real-world scenario~\cite{fan2018unsupervised} due to the non-trivial gap between training data and target domain. On the other hand, it is rather strenuous and impractical to annotate the target data, especially for online surveillance videos of large-scale camera network~\cite{lv2018unsupervised}. To directly make full use of the massive and cheap unlabeled video data, person Re-ID by unsupervised learning, where per camera-pair ID labeled training data is no longer required, is gaining increasing popularity~\cite{kodirov2015dictionary,wang2016towards,kodirov2016person,zhao2017person,Liu_2017_ICCV,Ye_2017_ICCV,ma2017person}. %However, most of them suffer limited performance. 

\begin{figure*}[t]
  \begin{center}
     \includegraphics[width=1.0\linewidth]{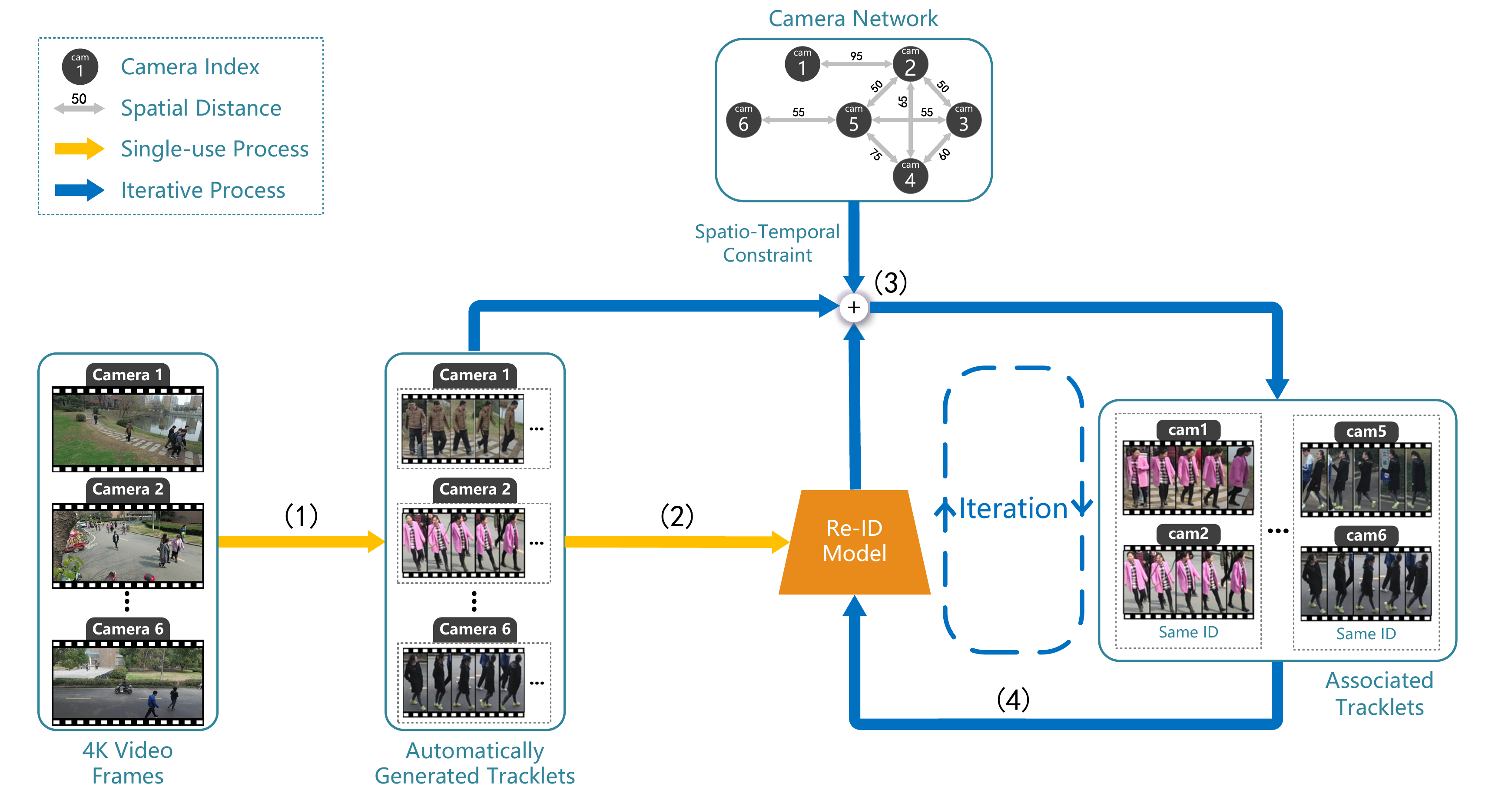}
  \end{center}
     \caption{Our framework consists of 4 steps: (1) Multi-person detection and tracking. (2) Within-camera Re-ID learning. (3) Cross-camera tracklets association with spatio-temporal regularization. (4) Cross-camera Re-ID learning. Step (1) and (2) only conduct once while (3) and (4) are iterated several times for progressive optimization.}
  \label{fig:framework}
\end{figure*}

In person Re-ID task, spatio-temporal context in camera network provides a wealth of information to distinguish as well as associate persons of interest~\cite{huang2016camera,martinel2017person,cho2019joint,lv2018unsupervised,wang2018spatial}. Since for most people the spatial transfer time between cameras usually follows a similar pattern of walking speed, it helps a lot in short-time person retrieval by eliminating lots of irrelevant images as the camera network is fixed. %However, it is still very challenging to deploy this method to real-world scenarios where people may frequently appear in the field of the camera network anytime from anywhere in an open environment. 

To learn a person Re-ID model, substantial cross-view training data is needed in order to cope with the significant visual appearance change between different cameras. To this end, cross-camera person tracklet association is an alternative to provide cross-view data for unsupervised Re-ID learning. As spatio-temporal information is beneficial to essentially improve the performance of short-time person Re-ID, it can also be explored in unsupervised cross-camera tracklet association. Considering the fact that it is more important to improve the precision of tracklet association for unsupervised methods than to sort out all the tracklets belonging to a person, it suffices to design a relatively simple tracklet association based on spatio-temporal constraint without any labeled data, even if we miss some positive tracklet pairs (with the same ID).

Based on the above motivation, in this paper, we propose a progressive unsupervised learning framework by cross-camera Tracklet Association with Spatio-Temporal Regularization (TASTR). Some previous unsupervised methods~\cite{lv2018unsupervised,Liu_2017_ICCV,wu2018exploit} either suffer the lack of accurate spatio-temporal information, or directly use off-the-shelf tracklets in training set for model initialization and assume different tracklets contain different person identities. To avoid the above issues and well justify the proposed approach, we have collected a new 4K UHD video dataset for unsupervised person Re-ID, which contains full frames and full spatio-temporal information. The general framework of the proposed approach is illustrated in Fig.~\ref{fig:framework}. In the first stage, we conduct multi-person detection and tracking per camera to get their tracklets and train a within-camera Re-ID model. In the second stage, cross-camera tracklets association with spatio-temporal regularization is proposed to obtain accurate pseudo labels across cameras to further learn cross-view identity-specific discriminative information. Finally, the Re-ID model is progressively optimized by iterating the two steps in the second stage. 

Our contributions can be summarized into three aspects:

\begin{itemize}%[noitemsep,nolistsep]
   \item We collect a new 4K UHD video dataset, named Campus-4K, for unsupervised person Re-ID. Compared with existing Re-ID datasets, Campus-4K is of higher quality with full frames and complete spatio-temporal information. 
   \item We propose a progressive unsupervised deep learning framework for person Re-ID by Tracklet Association with Spatio-Temporal Regularization (TASTR), which significantly improves the precision and recall rate of cross-camera tracklet association as well as the performance of the Re-ID model. 
   \item Extensive experiments show that our method notably improves the performance of unsupervised Re-ID. Our unsupervised method with spatio-temporal clues only in the training phase achieves competitive performance (rank-1: 76.4\% on Campus-4K and 74.1\% on DukeMTMC) with fully supervised methods (rank-1: 85.2\% on Campus-4K and 78.1\% on DukeMTMC) using the same batch hard triplet loss~\cite{hermans2017defense}.
\end{itemize}

\begin{figure*}
	\begin{center}
		\includegraphics[width=1.0\linewidth]{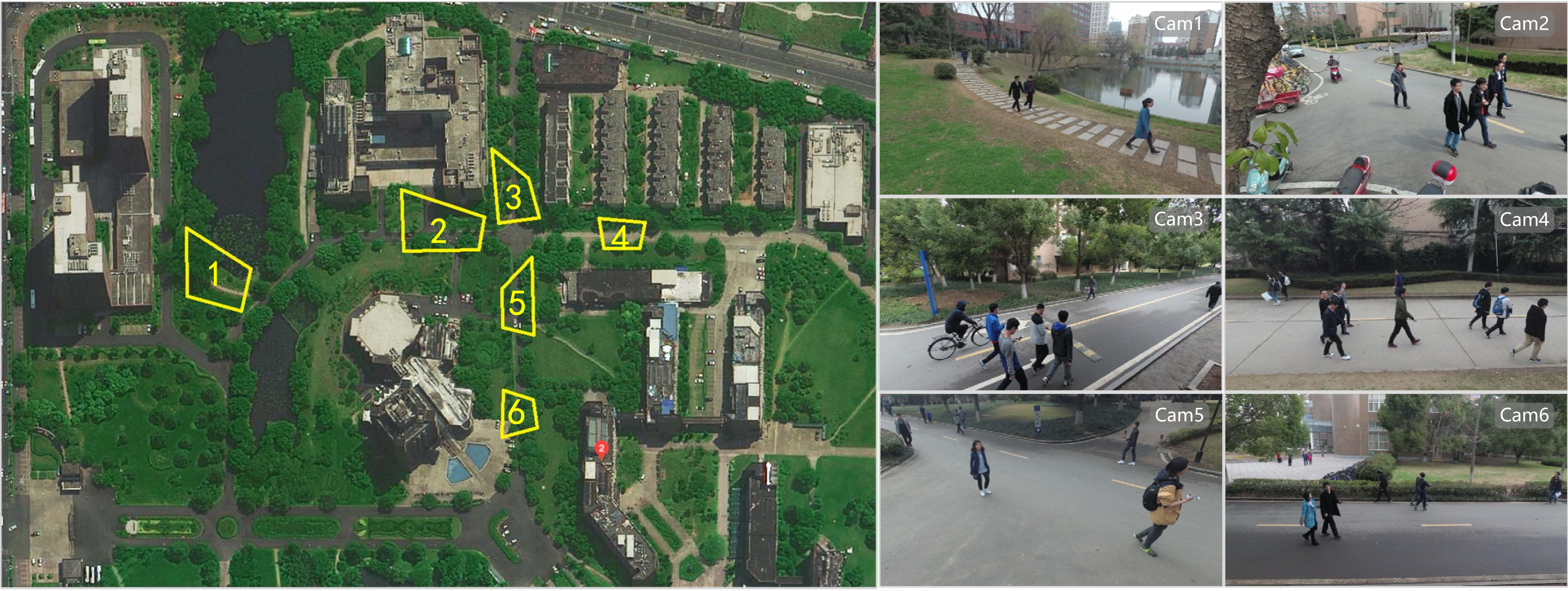}
	\end{center}
	\caption{Camera network of Campus-4K: 6 static non-overlapping synchronized
		3840$\times$2160 UHD cameras}
	\label{fig:network}
\end{figure*}

\section{Related Work}

In this section, we mainly review the unsupervised person Re-ID methods that are most related to the proposed approach, where pairwise ID labeled training data for each pair of camera views is not required in model learning.

%Unsupervised person Re-ID means no pairwise ID labeled training data for each pair of camera views is available while Sometimes the auxiliary source Re-ID dataset is not necessarily unlabelled~\cite{peng2016unsupervised,fan2018unsupervised,su2016deep,wang2018transferable}. 
%In this section, we mainly review the unsupervised person Re-ID methods that are most related to the proposed approach.

Early unsupervised person Re-ID methods mainly focus on feature representation learning~\cite{farenzena2010person,ma2012bicov,ma2012local}. Dictionary learning~\cite{kodirov2015dictionary,liu2014semi} and salience learning~\cite{zhao2017person,zhao2013unsupervised} methods are also proposed to learning salient and view-invariant representations. To balance the scalability and accuracy of the Re-ID model, semi-supervised learning~\cite{liu2014semi,wang2016towards} are proposed but they still assume sufficient cross-view labeled data for model training. Recently, some cross-dataset transfer learning methods~\cite{peng2016unsupervised,zhu2017unpaired,fan2018unsupervised,su2016deep,wang2018transferable,deng2018image} have been proposed to leverage the labeled data in other datasets to improve the performance on target dataset. These methods gain much better accuracy than the classical unsupervised methods, but they need the auxiliary source Re-ID dataset and still require latent similarity between the source domain and the unlabeled target domain.

Similar to our work, cross-camera tracklet association (labeling)~\cite{Liu_2017_ICCV,wu2018exploit,li2018unsupervised} is more scalable for unsupervised Re-ID with no extra data or assumption on the similarity between source and target domains. Due to the limitation of existing datasets, most of them do not perform Re-ID learning in a pure unsupervised way. Instead, they take for granted the tracklets provided by the dataset, which essentially assumes the perfect tracking in videos. However, such an assumption is somewhat too strong in real-world situations. Li \emph{et al.}~\cite{li2018unsupervised} propose a Tracklet Association Unsupervised Deep Learning (TAUDL) model to consider a pure unsupervised person Re-ID problem. They perform Sparse Space-Time Tracklet (SSTT) sampling for within-view tracklet labeling and formulate a Cross-Camera Tracklet Association (CCTA) loss for coarse-grained underlying cross-view tracklet association. However, SSTT throws away a lot of tracklets which may be useful for Re-ID learning while CCTA loss does not make explicit cross-view tracklet association. In contrast, our method can make full use of tracklet data and achieve accurate cross-camera tracklet association result, which leads to a considerable improvement in Re-ID performance.

% Spatio-temporal clues can promote short-time Re-ID precision significantly. But the spatio-temporal distribution in a real-world camera network may be very complicated because there may exist several paths between a camera pair and leads to several peaks in spatio-temporal distribution. Wang \etal~\cite{wang2018spatial} use a Histogram-Parzen method (HP) to model the probability of positive image pairs with respect to time difference for each camera pair. This method needs sufficient labeled data for distribution estimation and will miss some hard samples inevitably. However, 

Different from most existing unsupervised person Re-ID approaches, we perform a pure unsupervised person Re-ID at the very beginning (multi-person detection and tracking) without assuming that the tracklets per camera are off-the-shelf and different tracklets indicate different person identities. Actually, the tracklets can be interrupted and the number of tracklets is more than the number of identities per camera. Experiments show that person tracking not only provides within-camera training data for Re-ID but also benefits from within-camera person discriminative learning. Then with the help of spatio-temporal constraint, we considerably promote the precision and recall rate of cross-camera tracklet association. Finally, we iterate cross-camera tracklet association and Re-ID learning to optimize the Re-ID model progressively and achieve competitive performance with supervised methods using the same triplet loss.

\section{Campus-4K Dataset}

\begin{figure*}
   \begin{center}
      \subfigure[]{
         \includegraphics[width=0.39\linewidth]{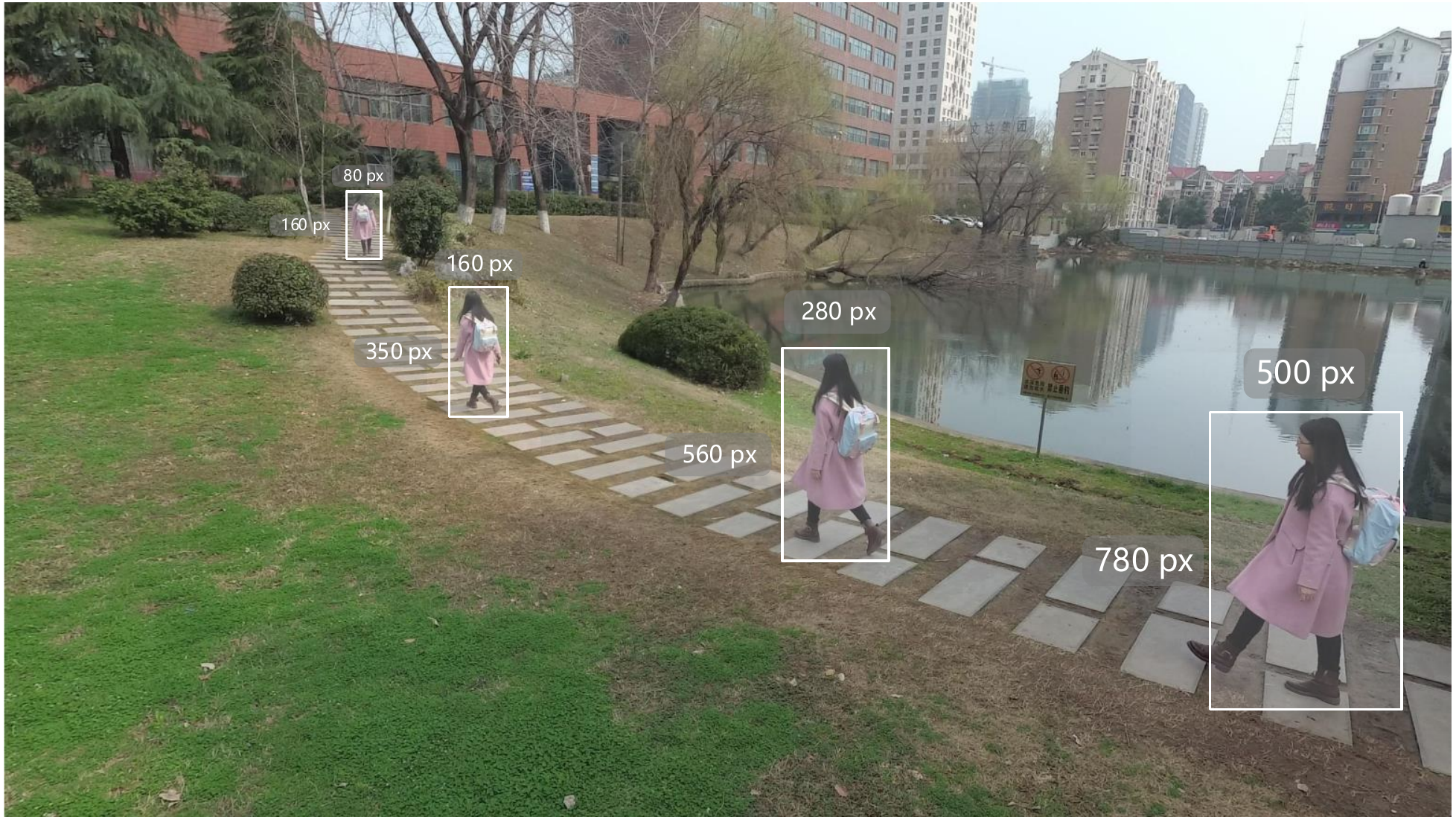}\label{fig:scale-1}
      }
      \subfigure[]{
         \includegraphics[width=0.57\linewidth]{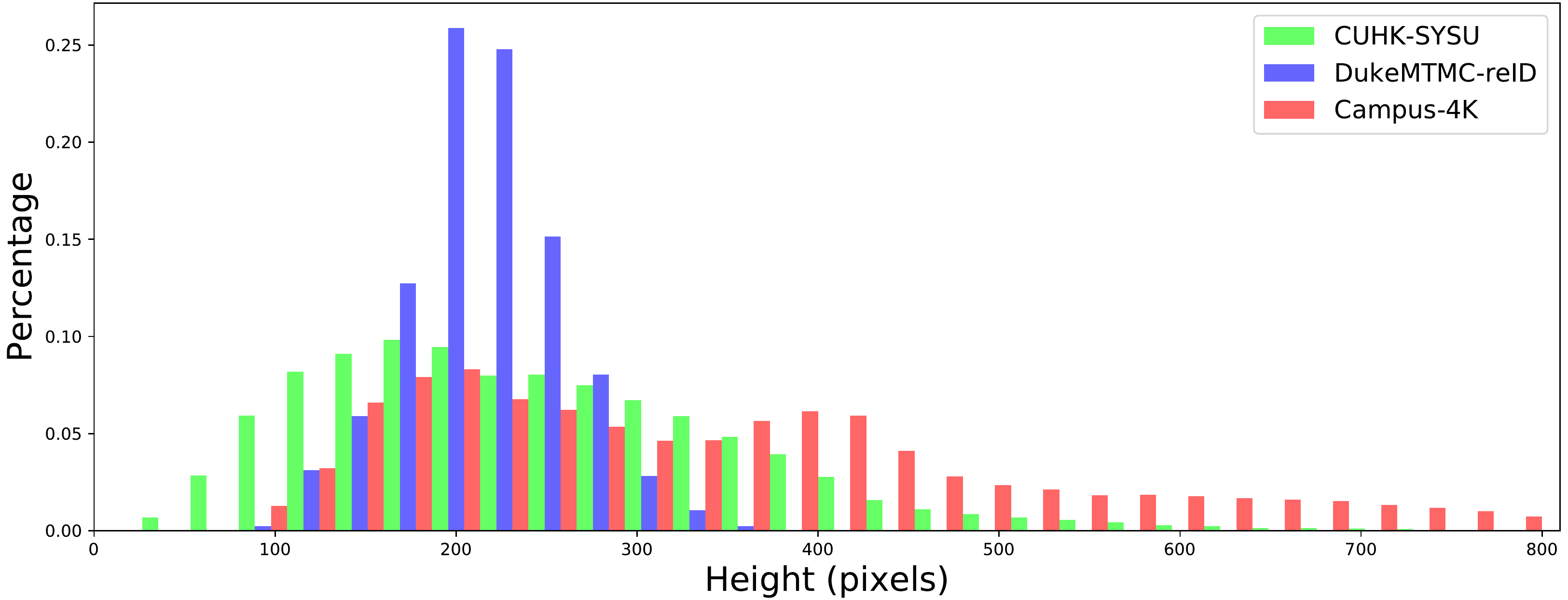}\label{fig:scale-2}
      }
   \end{center}
   \caption{(a) Illustration of person scale variance in our Campus-4K dataset. (b) The scale (height) distributions of person bounding boxes in three datasets. It is notable that the person's scale in Campus-4K covers a much diverse range than the others.}
\end{figure*}

\subsection{Dataset Description}

Thanks to many public Re-ID datasets~\cite{gray2008viewpoint,hirzer2011person,li2014deepreid,zheng2015scalable,wang2016person,zheng2017unlabeled,wei2018person}, great progress has been made for person Re-ID in recent years. In most existing Re-ID datasets, the resolution is low and the person images are usually blurred with indistinguishable characteristic detail, which imposes significant difficulty on the algorithm design. To this end, we have collected a new 4K UHD video dataset, \emph{i.e.}, Campus-4K\footnote{\url{http://home.ustc.edu.cn/~xieqiaok/campus-4k}}, with full frames and full spatio-temporal information. It is collected by a non-overlapping outdoor camera network on campus (see Fig.~\ref{fig:network}), which consists of six synchronized 3840$\times$2160 UHD cameras. It aims to provide a high-quality Re-ID dataset for both supervised and unsupervised Re-ID, and in this paper we mainly focus on unsupervised learning. 

\begin{table}
	\begin{center}
		\resizebox{0.38\textwidth}{!}{
			\begin{tabular}{c||c|c|c}
				\hline
				Camera & \# Identities & \# Tracklets & \# BBoxes \\
				\hline\hline
				Cam 1 & 331 & 335 & 59,021 \\
				Cam 2 & 649 & 660 & 147,679 \\
				Cam 3 & 645 & 653 & 83,569 \\
				Cam 4 & 841 & 843 & 62,307 \\
				Cam 5 & 689 & 713 & 123,230 \\
				Cam 6 & 643 & 645 & 45,503 \\
				\hline
				Total & 1,567 & 3,849 & 521,309\\
				\hline
			\end{tabular}
		}
	\end{center}
	\caption{Statistics of Campus-4K.}\label{tab:statistics-1}
\end{table}

\begin{table}
	\begin{center}
		\resizebox{0.38\textwidth}{!}{
			\begin{tabular}{c||c|c|c}
				\hline
				Subset & \# Identities & \# Tracklets & \# BBoxes \\
				\hline\hline
				Train & 695 & 1,843 & 243,998 \\
				Query & 695 & 695 & 97,914 \\
				Gallery & 872 & 1,311 & 179,397 \\
				\hline
			\end{tabular}
		}
	\end{center}
	\caption{Evaluation setting of Campus-4K.}\label{tab:statistics-2}
\end{table}

Our Campus-4K dataset carries the spatio-temporal context information about the camera position and recorded videos. In some recent work~\cite{wang2018spatial,lv2018unsupervised}, it has been shown that spatio-temporal context is of great importance to improve the Re-ID accuracy. Such information is also easily accessible in practice. 

%There still remains a big gap between the performance of unsupervised and supervised methods. However, some information is ignored but has great potential for improving Re-ID performance. As shown in some recent researches~\cite{wang2018spatial,lv2018unsupervised}, spatio-temporal context is very useful and easy to obtain in practice. It is naturally to record the time while capturing video, and the position of all cameras is also directly available. Campus-4K is such a dataset that you can get the people's images with their locations and time stamps simultaneously.

% Previously, due to the limitations of datasets, most unsupervised Re-ID methods could only use images themselves across different cameras to guide Re-ID learning. And 

To our best knowledge, Campus-4K is the first public 4K (3840$\times$2160) UHD dataset for person Re-ID. In addition, full frames and full spatio-temporal information are provided. In recent years, there has been a trend of increasing resolutions for surveillance cameras. As video quality improves, some fine-grained details of people such as textures of clothes, carry-on items and even face become available. They make it easier to recognize many attributes that were previously difficult to classify, and it also brings new challenges such as large scale (resolution) variance of person bounding boxes, and the joint combination of face recognition and person Re-ID in the wild. 

As shown in Fig.~\ref{fig:scale-1}, since 4K cameras have a broader view than ordinary cameras and the distance between person and camera is uncontrolled, the scale of people may vary significantly, which introduces a multi-scale matching problem~\cite{lan2018person}. As shown in Fig.~\ref{fig:scale-2}, in Campus-4K, most person bounding boxes are in the range of about 100 to 800 pixels in height, and it covers a more diverse range than other existing datasets. Moreover, the face regions in Campus-4K are of relatively high resolution, which makes it possible to perform face recognition tasks. However, since the faces are not always visible, and are taken in different views, lighting conditions and scales, it is still very challenging to link face recognition to person Re-ID, which is out of scope for this work. 

\subsection{Evaluation Protocol}

The time of all cameras is synchronized, and $77,195$ 4K frames are collected continuously at 30 fps per camera. As ground truth, person bounding boxes with identity information are generated by manual annotation with the help of tracking and temporal smoothing. We split our dataset into training set and testing set according to the moment that the person was first captured by our camera network. For all identities who appear in at least two camera views, we denote by $n$ the index of the earliest frame of his/her appearance in all camera views. Identities with $n$ less than 46,410 would be assigned to the training set, and the rest would be assigned to the testing set. Besides, identities who appear in only one camera would be considered as distractors. 

During testing, one tracklet of each person in testing set would be selected to form query set, and the gallery set consists of the rest tracklets in testing set along with all distractors to make the evaluation more challenging. Images of each video are sampled every three frames (0.1 seconds) and there are $135$ images per person per video on average. More details can be found in Table~\ref{tab:statistics-1} and Table~\ref{tab:statistics-2}. It should be noted that if full frames are needed for unsupervised Re-ID, only frames whose indices $n$ are less than $46,410$ can be used for training to avoid the use of identities from the testing set.

Like most existing Person Re-ID datasets, Cumulated Matching Characteristics (CMC) curve is used to evaluate the performance of Re-ID methods. For each query video, multiple ground truths may exist and the CMC curve may be biased since ``recall" is not considered~\cite{zheng2015scalable}. Therefore, mean Average Precision (mAP) is also used for overall performance evaluation.

\section{The Proposed Method}

%\subsection{Framework Overview}

A practical person Re-ID system in surveillance usually consists of three modules, \emph{i.e.}, person detection, tracking and re-identification~\cite{zheng2016person}. Although they are generally considered as three independent computer vision tasks, they can complement each other to improve performance. In this paper, we propose a novel progressive unsupervised Re-ID approach by Tracklet Association with Spatio-Temporal Regularization (TASTR). The proposed framework is shown in Fig.~\ref{fig:framework}, which consists of four key steps as follows. First, for each camera, we obtain tracklets by multi-person detection and tracking for initialization. Then, we conduct within-camera training for person Re-ID based on the initially obtained tracklets. After that, we associate different person tracklets across cameras with Spatio-Temporal Regularization (STR). Finally, based on the cross-camera tracklet association results, we re-train the Re-ID model to generate better feature representation. %The last three stages can be iterated to boost the performance. 

The first two steps focus on within camera exploration, and generate within-camera Re-ID model. The remaining two steps are iterated multiple times as the Re-ID model may generate more and better matching pairs, which can be used for more effective Re-ID learning. In other words, we optimize the Re-ID model in a progressive fashion. 

\subsection{Multi-person Detection and Tracking}\label{sec:step-1}

\begin{figure*}
	\begin{center}
		\subfigure[YOLOv3]{
			\includegraphics[width=0.48\linewidth]{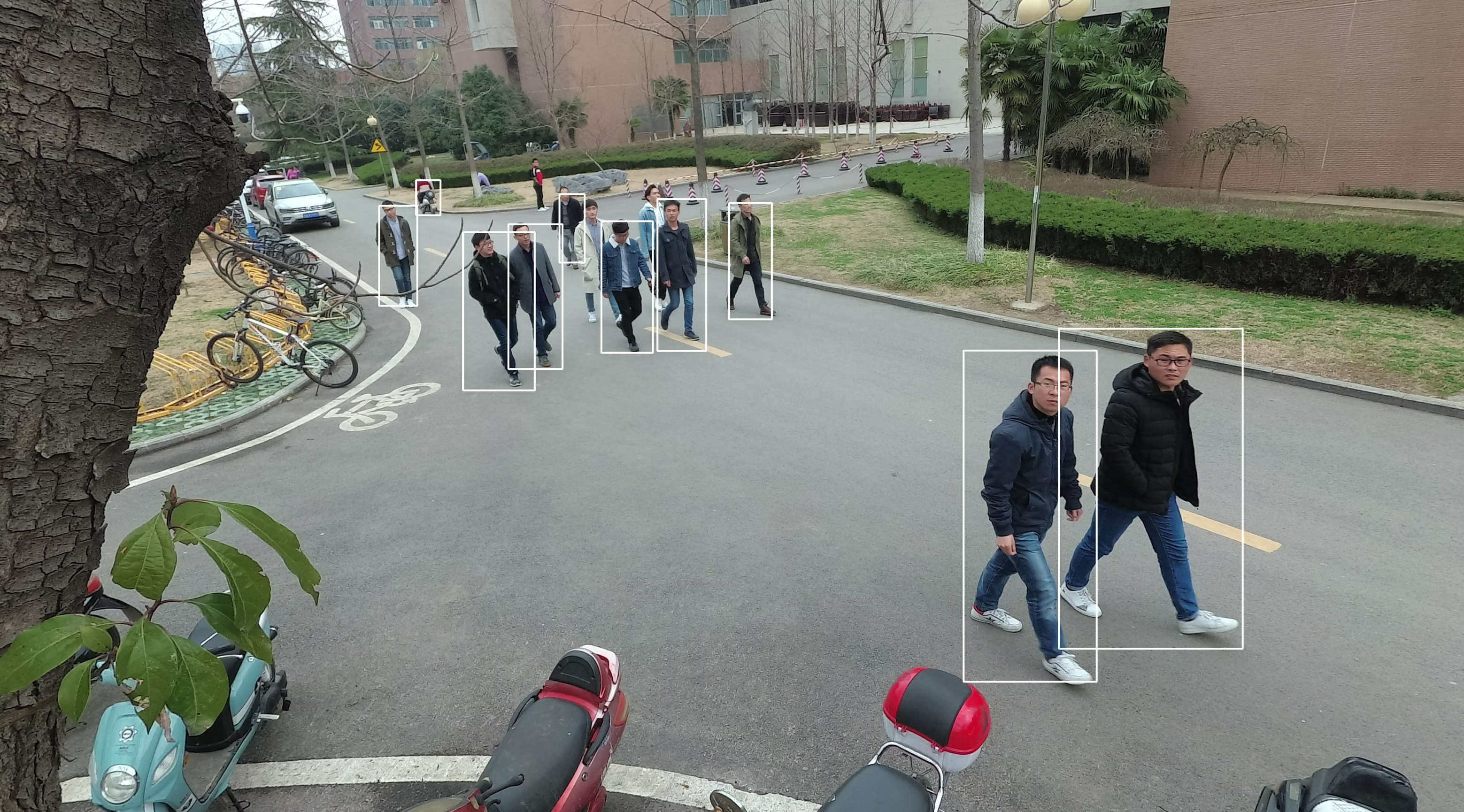}\label{fig:scale-1}
		}
		\subfigure[AlphaPose]{
			\includegraphics[width=0.48\linewidth]{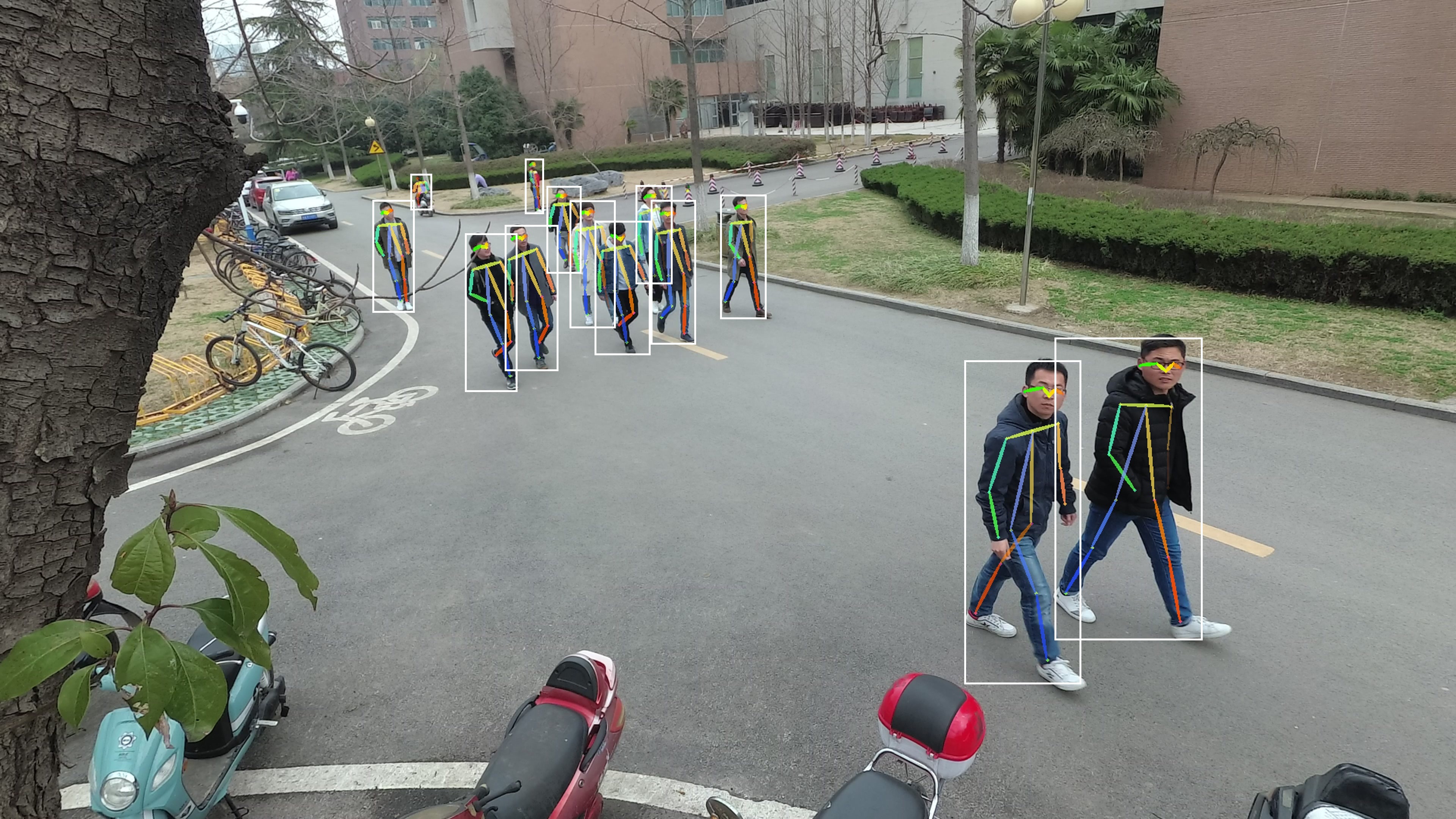}\label{fig:scale-2}
		}
	\end{center}
	\caption{Examples of multi-person detection result \emph{w.r.t.} different detection methods: (a) YOLOv3, (b) AlphaPose. YOLOv3 suffers from various problems such as detection error, location error and miss detection while AlphaPose can provide much more stable and fine-grained results. Better detections are also beneficial for subsequent multi-person tracking. And AlphaPose with PoseFlow can automatically generate high quality data for unsupervised perosn Re-ID learning initialization.}
	\label{fig:pose}
\end{figure*}

%\begin{figure}
%   \begin{center}
%      \includegraphics[width=0.98\linewidth]{figures/pose.jpg}
%   \end{center}
%   \caption{Example of multi-person pose estimation and the exacted patches of person derived from these poses.}\label{fig:pose}
%\end{figure}

We exploit person detection and tracking to generate initial data for unsupervised person Re-ID learning. To begin with, we detect all persons in each frame by AlphaPose~\cite{fang2017rmpe}. AlphaPose generates high-quality pose estimation from person bounding boxes obtained by detector, \emph{e.g.}, YOLO~\cite{redmon2016you}, and is resilient against imperfect detection. Then, we track the pose each person with PoseFlow~\cite{xiu2018pose}, which associates cross-frame poses stably with an online optimization method. The bounding box of each person is derived from the pose (see Fig.~\ref{fig:scale-2}), which is used to crop the image patch for the following tracklet association and Re-ID learning. 

It is an alternative to use YOLO (or Faster-RCNN~\cite{ren2015faster}) for detection and IOU for tracking. However, as shown in Fig.~\ref{fig:scale-1} the general object detection methods such as YOLO and Faster-RCNN are unable to extract fine-grained human bodies and suffer from instability in detecting people of various scales. Thus, an additional learning-based alignment is often required to refine the person detection results to benefit the following person Re-ID task. In contrast, based on human pose estimation with AlphaPose and pose tracking with PoseFlow, we can obtain more stable and fine-grained results, which is free of person alignment for person Re-ID.

%For a pure unsupervised Re-ID methods, multi-person detection and tracking can provide rich raw data for subsequent Re-ID learning. We adopt AlphaPose~\cite{fang2017rmpe} for pose estimation, then perform PoseFlow~\cite{xiu2018pose} for pose tracking. It associates cross-frame poses by an online optimization methods to form pose flows (PF-Builder), then design a pose flow NMS (PF-NMS) to robustly reduce needless pose flows and re-link temporal disjoint ones. The combination of AlphaPose and PoseFlow can achieve good multi-person tracking performance.

\subsection{Within-camera Re-ID Learning}\label{sec:step-2}

\begin{figure*}
   \begin{center}
      \includegraphics[width=0.95\linewidth]{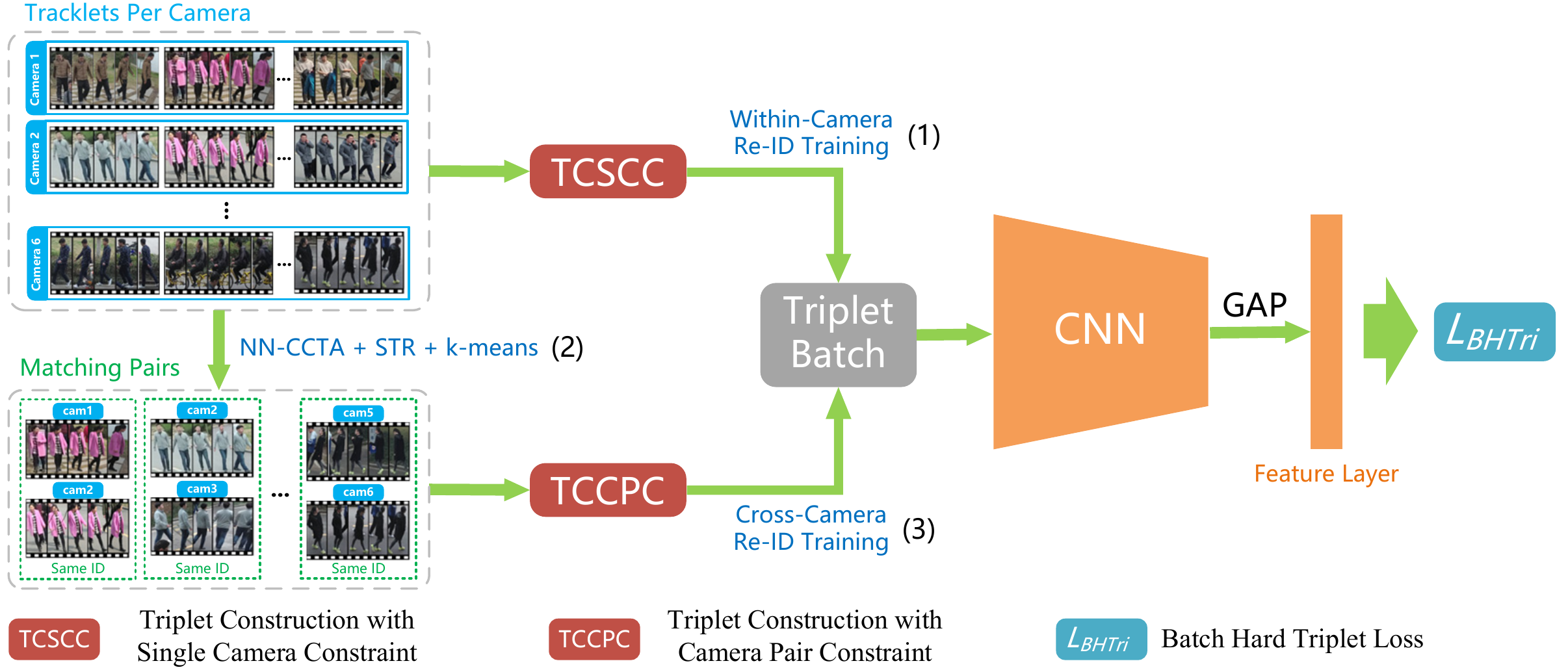}
   \end{center}
      \caption{The pipeline of Re-ID training. (1) Given tracklets per camera, TCSCC sampling strategy is used to form the triplet batch for within-camera Re-ID learning. (2) Based on the learned model we can perform 1-reciprocal Nearest Neighbor based Cross-Camera Tracklet Association (NN-CCTA) with Saptio-Temporal Regularization (STR) and $k$-means to obtain accurate matching tracklets for each camera pair. (3) TCCPC is used to form triplet batch for cross-camera Re-ID learning. In this figure, process (1) is performed only once while processes (2) and (3) are iterated for progressive optimization.}
   \label{fig:tastr}
\end{figure*}

With AlphaPose and PoseFlow, we obtain a series of tracklets each of which contains the temporal patches of a certain person. However, due to imperfect tracking, it is inevitable that the trajectory of a person will be fragmented into different tracklets, especially when people are occluded. 
Due to the limitation of datasets, many traditional unsupervised methods directly use complete tracklets given by training data and assume different tracklets indicate different identities. Such a setting ignored the tracklet identity duplication problem, \emph{i.e.}, different tracklets may correspond to the same person ID. 

Observing that in surveillance videos such as people re-appearing in a camera view is rare during a short time period and most people travel through a single camera view in a common time period $Q<T$, Li \emph{et al.}~\cite{li2018unsupervised} propose a Sparse Space-Time Tracklet (SSTT) sampling method and treat each camera view separately as a classification task. However, it risks declining many tracklets that may be helpful for Re-ID learning. 
Instead of considering it as a classification problem, we adopt triplet loss for Re-ID training. To avoid losing too many tracklets that are potentially useful for Re-ID learning, we use all tracklets to perform within-camera Re-ID training. 
%However, it will bring about two problems: cross-camera ID duplication due to the lack of cross-view ID labels and within-camera ID duplication caused by imperfect tracking. 

Concretely, we adopt triplet loss with hard mining proposed by Hermans \emph{et al.}~\cite{hermans2017defense}. For each triplet batch, we randomly sample $P$ tracklets, and then randomly sample $K$ images for each tracklet, resulting in a batch of $P \times K$ images. To deal with false negatives caused by ID duplication as much as possible, we proposed a sampling strategy called \emph{Triplet Construction with Single Camera Constraint} (TCSCC). It requires that all $P$ sampled tracklets within the batch are from the same camera and the time interval between them must be greater than a gap of $T$. Different from SSTT~\cite{li2018unsupervised}, we do not decline tracklets that do not satisfy the above conditions because they may appear in other triplet batches as well. Finally, for each anchor $\mathbf{a}$ in the batch, only the hardest positive and negative samples will be considered for computing the loss, which is a build-in hard sample mining method called \emph{Batch Hard}. The batch hard triplet loss is computed as:

\begin{equation}
    \begin{split}
    \mathcal{L}_{BHTri} = \overbrace{\sum\limits_{i=1}^{P} \sum\limits_{a=1}^{K}}^{\textnormal{all anchors}}
    \Big[ m & + \overbrace{\max\limits_{p=1 \dots K} D\left(f(x^i_a), f(x^i_p)\right)}^{\textnormal{hardest positive}}\\
    & - \underbrace{\min\limits_{\substack{j=1 \dots P \\ n=1 \dots K \\ j \neq i}} D\left(f(x^i_a), f(x^j_n)\right)}_{\textnormal{hardest negative}} \Big]_+,
    \end{split}
\end{equation}
where $m$ is the margin of triplet loss, $x^i_j$ corresponds to the $j$-th image of the $i$-th person in the batch, $f(x^i_j)$ denotes its feature, $D(\cdot,\cdot)$ means distance, and $[\cdot]_+ = max(0,\cdot)$.

With the above triplet loss based learning, we can obtain a preliminary Re-ID model, denoted as TASTR-S1, with strong within-camera ID discrimination capability (see Sec.~\ref{sec:s1} and Fig.~\ref{fig:s1}), which, in turn, is helpful for within-camera tracklets association (re-link). In other words, the Re-ID model learns within-view ID discriminative information from tracking and can improve tracking performance in return.

\subsection{Cross-Camera Tracklets Association}\label{sec:step-3}

There is still a big gap between TASTR-S1 and the supervised model due to the lack of cross-view pairwise data. To address this problem, we need to associate cross-camera tracklets for global ID discriminative learning. It may not be satisfactory to directly use TASTR-S1 for cross-view tracklet association as it may be weak in associating the same person in cross-view circumstances. However, some extra clues like spatio-temporal information which is easily available in real-world practices has great potential for improving the precision and recall of tracklet association. 

It is often difficult to model the spatio-temporal pattern of moving people because of the diverse paths and uncontrollable pedestrians among different cameras. However, in unsupervised Re-ID training, it is more important to obtain correctly matched cross-view tracklets that belong to the same ID for cross-view ID discriminative learning than sorting out all the tracklets of a person. 
%Therefore, we make use of the assumption that most people move with definite purposes at similar speeds, as well as the camera network is fixed.
We make use of the assumption that most people move with definite purposes at similar speeds, as well as the camera network is fixed. So spatio-temporal constraint can eliminate plenty of irrelevant images and narrow the search space for cross-camera tracklet association. This can considerably improve the precision and quantity of associated tracklets with limited hard samples missing. 

Formally, given two tracklets $T_i$ and $T_j$ ($i$ and $j$ denote tracklet indexes), we extract visual features by TASTR-S1 and get two feature vectors $\boldsymbol x_i$ and $\boldsymbol x_j$, respectively. Then we compute the Euclidean distance between them:
\begin{equation}
   D( T_i, T_j ) = \| \boldsymbol x_i - \boldsymbol x_j \|.
\end{equation}

On the other hand, we use a simple but effective Gaussian function to reflect the spatio-temporal constraint for a camera pair:
\begin{equation}
   R(\Delta t, c_i, c_j) = exp\Big(-\frac{(\Delta t - \overline{t}_{c_i,c_j})^2}{2 \sigma_{c_i,c_j}}\Big),
\end{equation}
where $\Delta t$ denotes the time interval between $T_i$ and $T_j$, $c_i$ means the camera index of the $i$-th tracklet, $\overline{t}_{c_i,c_j}$ is the average moving time between the camera pair ($c_i, c_j$), and $\sigma_{c_i,c_j}$ is its standard deviation. Without labeled training data, $\overline{t}_{c_i,c_j}$ is estimated by the spatial path length between the camera pair and pedestrians' average speed, and we set $\sigma_{c_i,c_j}=\lambda \overline{t}_{c_i,c_j}$. Therefore, farther distance result in larger $\overline{t}_{c_i,c_j}$ and $\sigma_{c_i,c_j}$

Finally, we compute the joint distance by visual feature and Spatio-Temporal Regularization (STR) as follows,
\begin{equation}
   \label{eq:djoint}
   D_{\text{\emph{joint}}}( T_i, T_j ) = \frac{D( T_i, T_j )}{R(\Delta t, c_i, c_j)}.
\end{equation}

From Eq.~(\ref{eq:djoint}), we can see that if the time interval of a tracklet pair is far from the average transfer time of the corresponding camera pair, it will lead to a small regularization item $R$, so the $D_{\text{\emph{joint}}}$ would increase as spatio-temporal prior indicates it is unlikely to have the same identity. In this way, some abnormal true matching pairs may be suppressed. However, much more true matches with common transfer time can be obtained since it eliminates lots of irrelevant matching tracklets and thus narrows the search space of cross-camera tracklet association. It can effectively improve the accuracy and recall rate of matching pairs~(\emph{i.e.} more quantity and higher quality), which is quite important for improving the performance of unsupervised Re-ID training.

Then we perform cross-camera tracklet association based on $D_{\text{\emph{joint}}}$. Specifically, for each camera pair, given a tracklet in one camera as probe and all tracklets in another camera as gallery, the ranking list can be computed by their joint distance. If we directly use the top match as the association result, it can lead to many false matches. As $k$-reciprocal nearest neighbors are more likely to be relevant to the probe~\cite{zhong2017re}, we adopt 1-reciprocal Nearest Neighbor based Cross-Camera Tracklet Association (NN-CCTA) as a strong constraint to identify possible matched candidates, \emph{i.e.}, both of them are the top one match of each other. 

After all the cross-camera tracklet pair candidates are obtained, we perform $k$-means for further refining, which is important for progressive improvements. Specifically, it is 1-D $k$-means on the Euclidean distances of all tracklet pair candidates for each camera pair, and the $k$ initial points are evenly located between the minimum and maximum distance values. Finally, the tracklet pairs that belong to the cluster with minimum average distance are selected as matching pairs for cross-view Re-ID training. 

\subsection{Cross-camera Re-ID Training}\label{sec:step-4}

So far we have got many cross-view tracklet pairs with high confidence via strict matching. Similar to Sec.~\ref{sec:step-2}, we use batch hard triplet loss for cross-camera Re-ID training. All matching tracklets of all camera pairs are adopted for cross-view discriminative learning, but in each triplet batch only those tracklets from the same camera pair are sampled to deal with ID duplication problem between different camera pairs. This sampling strategy is denoted as \emph{Triplet Construction with Camera Pair Constraint} (TCCPC). 

The pipeline of Re-ID learning is shown in Fig.~\ref{fig:tastr}. With the help of cross-camera tracklet association, the Re-ID model can achieve stronger cross-view ID discrimination capability. In other words, based on the refined model, we extract more discriminative visual feature. With the better visual feature and the spatio-temporal clues, we can derive higher quality cross-view  matching pairs by cross-camera tracklet association. Based on such motivation, we repeat cross-camera tracklet association and Re-ID training for several times. As a result, the Re-ID model can get progressive improvements in a mutual promotion manner.

\section{Experiments}
\subsection{Setup}
\subsubsection{Datasets}
Most existing Re-ID datasets lack both synchronized time-stamp and spatial distribution information of the camera network. Market1501~\cite{zheng2015scalable} and GRID~\cite{loy2009multi} provide the frame numbers in video sequences, which can be used as time-stamps, but it is unknown whether the time is synchronized and the location and coverage \emph{w.r.t.} each camera is not given as well. DukeMTMC-reID~\cite{zheng2017unlabeled} also provides the frame numbers, and it is a subset of the multi-target multi-camera tracking dataset DukeMTMC~\cite{ristani2016performance}. The cameras of DukeMTMC are synchronized and the spatial distribution of the cameras is also provided. So besides our Campus-4K dataset, we choose DukeMTMC-reID to evaluate the proposed unsupervised TASTR model. No spatio-temporal information is used in testing phase on both datasets.

Campus-4K is a new Re-ID dataset with full frames and full spatio-temporal information. It is readily ready for the evaluation of the unsupervised Re-ID learning from multi-person detection and tracking. For DukeMTMC-reID, as the cropped person images are off-the-shelf, we consider the image patches of a person in one camera as a tracklet and ignore its label for unsupervised learning as most previous unsupervised methods do. It is notable that this is not a pure unsupervised setting as there is no within-camera ID duplication problem so different tracklets per camera contain different person identities. 

\subsubsection{Implementation Details}
We use the ResNet-50~\cite{he2016deep} model pre-trained on ImageNet as the backbone, and train the model with batch hard triplet loss. Images are resized to 256$\times$128, and we augment the training images online with cropping, horizontal flip and normalization. To balance the number of images of different tracklets as some tracklets may contain hundreds of images, at most 60 images of each tracklet would be randomly selected for training. We use Adam~\cite{kingma2014adam} as the optimizer and the initial learning rate is 0.0003. We set $k=3$ for $k$-means, $\lambda=0.7$ for STR, and the number of iterations is fixed to 5. Our code and models are available at \url{https://github.com/xieqk/TASTR}.

\subsection{Comparison to the State-of-the-Art Methods}

\begin{table}
	\footnotesize
	\begin{center}
		%		\resizebox{0.47\textwidth}{!}{%
		\begin{tabular}{c||c||ccc}
			\hline
			Methods & Reference & rank-1 & rank-5 & mAP \\
			\hline
			LOMO~\cite{liao2015person} & CVPR'15 & 12.3 & 21.3 & 4.8\\
			BOW~\cite{zheng2015scalable} & ICCV'15 & 17.1 & 28.8 & 8.3 \\
			UDML~\cite{peng2016unsupervised} & CVPR'16 & 18.5 & 31.4 & 7.3 \\
			\hline
			PUL$^\dagger$~\cite{fan2018unsupervised} & TOMM'18 & 30.4 & 44.5 & 16.4 \\
			CycleGAN$^\dagger$~\cite{zhu2017unpaired} & ICCV'17 & 38.5 & 54.6 & 19.9 \\
			SPGAN$^\dagger$~\cite{deng2018image} & CVPR'18 & 41.1 & 56.6 & 22.3 \\
			TJ-AIDL$^\dagger$~\cite{wang2018transferable} & CVPR'18 & 44.3 & 59.6 & 23.0 \\
			SPGAN+LMP$^\dagger$~\cite{deng2018image} & CVPR'18 & 46.9 & 62.6 & 26.4 \\
			HHL$^\dagger$~\cite{zhong2018generalizing} & ECCV'18 & 46.9 & 61.0 & 27.2 \\
			TAUDL~\cite{li2018unsupervised} & ECCV'18 & 61.7 & - & 43.5 \\
			UTAL~\cite{li2019unsupervised} & TPAMI'19 & 62.3 & - & 44.6 \\
			MAR~\cite{yu2019unsupervised} & CVPR'19 & 67.1 & 79.8 & 48.0 \\
			\hline
			\textbf{TASTR} & This work & \textbf{74.1} & \textbf{85.5} & \textbf{54.9} \\
			\hline
		\end{tabular}%}%
	\end{center}
	\caption{Comparison of the proposed TASTR method with state-of-the-art unsupervised methods on DukeMTMC-reID. ``-'' : No reported result or implementation code is available. $^\dagger$: Transfer based model using extra training data.}
	\label{tab:sota}
\end{table}

\begin{table*}
	%   \footnotesize
	\begin{center}
		\resizebox{1.0\textwidth}{!}{%
			\begin{tabular}{c|ccc|ccccc|ccccc}
				\hline
				\multirow{2}{*}{Group}
				& \multicolumn{3}{|c|}{Dataset} & \multicolumn{5}{|c|}{Campus-4K} & \multicolumn{5}{|c}{DukeMTMC-reID} \\
				\cline{2-14}
				& \multicolumn{3}{|c|}{Methods} & rank-1 & rank-5 & rank-10 & rank-20 & mAP & rank-1 & rank-5 & rank-10 & rank-20 & mAP \\
				\hline\hline
				\multirow{2}{*}{(1)}
				& \multicolumn{3}{|c|}{BHTri (Market1501)} & 37.3 & 57.6 & 65.8 & 74.1 & 39.3 & 24.1 & 39.4 & 46.6 & 54.2 & 12.1 \\
				& \multicolumn{3}{|c|}{BHTri (Target Dataset)} & 85.2 & 96.1 & 97.8 & 98.8 & 87.5 & 78.1 & 88.5 & 91.4 & 93.8 & 60.9 \\
				\hline\hline
				\multirow{2}{*}{(2)}
				& \multicolumn{3}{|c|}{TASTR-S1 w/o TCSCC} & 5.2 & 15.5 & 23.6 & 33.4 & 8.6 & 13.0 & 20.0 & 24.9 & 30.6 & 6.5 \\
				& \multicolumn{3}{|c|}{TASTR-S1} & 53.0 & 74.7 & 83.2 & 88.3 & 53.9 & 56.4 & 71.5 & 77.0 & 82.6 & 38.4 \\
				\hline\hline
				\multirow{9}{*}{(3)}
				& STR & $k$-means & * & rank-1 & rank-5 & rank-10 & rank-20 & mAP & rank-1 & rank-5 & rank-10 & rank-20 & mAP \\
				\cline{2-14}
				& & & & 63.4 & 82.3 & 88.1 & 92.9 & 65.3 & 67.6 & 81.3 & 86.1 & 88.5 & 47.2 \\
				& & & \ding{51} & 61.5 & 81.6 & 87.4 & 91.4 & 63.2 & 67.5 & 80.7 & 84.6 & 88.4 & 46.3 \\
				\cline{2-14}
				& \ding{51} & & & 63.9 & 82.9 & 88.2 & 93.1 & 65.3 & 69.4 & 82.0 & 85.9 & 89.4 & 48.3 \\
				& \ding{51} & & \ding{51} & 64.0 & 82.5 & 88.0 & 92.0 & 65.1 & 67.4 & 81.0 & 85.5 & 88.7 & 44.6 \\
				\cline{2-14}
				& & \ding{51} & & 53.9 & 73.0 & 79.4 & 84.9 & 53.7 & 57.4 & 73.2 & 78.9 & 83.1 & 37.4 \\
				& & \ding{51} & \ding{51} & 56.2 & 78.1 & 82.7 & 87.8 & 59.3 & 61.1 & 75.5 & 80.8 & 84.9 & 40.1 \\
				\cline{2-14}
				& \ding{51} & \ding{51} & & 65.3 & 83.6 & 88.9 & 92.4 & 65.8 & 69.3 & 81.8 & 86.1 & 89.1 & 48.1 \\
				& \ding{51} & \ding{51} & \ding{51} & \textbf{76.4} & \textbf{91.6} & \textbf{94.6} & \textbf{96.4} & \textbf{78.3} & \textbf{74.1} & \textbf{85.5} & \textbf{89.0} & \textbf{91.8} & \textbf{54.4} \\
				\hline
		\end{tabular}}%
	\end{center}
	\caption{Comparison of the effectiveness of different components. They are categorized into three groups. Group: (1) Supervised models trained on extra or target dataset. (2) Within-camera Re-ID training. (3) Cross-camera Re-ID training. STR: Spatio-Temporal Regularization, $^*$: Progressive optimization.}
	\label{tab:component}
\end{table*}

We compare our TASTR model with some existing unsupervised state-of-the-art methods on DukeMTMC-reID as shown in Table~\ref{tab:sota}. Among the comparison methods, BOW~\cite{zheng2015scalable}, LOMO~\cite{liao2015person} and UDML~\cite{peng2016unsupervised} are hand-crafted feature representation based methods. Besides, five other methods are based on transfer learning with extra training data like Market1501: PUL~\cite{fan2018unsupervised}, CycleGAN~\cite{zhu2017unpaired}, SPGAN~\cite{deng2018image}, TJ-AIDL~\cite{wang2018transferable} and HHL~\cite{zhong2018generalizing}. Then, TAUDL~\cite{li2018unsupervised} and UTAL~\cite{li2019unsupervised} are unsupervised tracklet association based methods, which are closest to our work. The last one, MAR~\cite{yu2019unsupervised} conducts soft multilabel learning for unsupervised Re-ID with the help of a set of known reference persons from an auxiliary Re-ID dataset.

From Table~\ref{tab:sota}, it is observed that BOW~\cite{zheng2015scalable}, LOMO~\cite{liao2015person} and UDML~\cite{peng2016unsupervised} achieve very limited performance, while transfer based model obtain higher performance than hand-crafted features, for instance, HHL~\cite{zhong2018generalizing} achieves 46.9\% rank-1 accuracy. Comparing with unsupervised transfer-based methods, tracklet association based unsupervised methods such as TAUDL~\cite{li2018unsupervised}, UTAL~\cite{li2019unsupervised} and the proposed TASTR are much superior, and TASTR outperforms all the competing methods. Notably, no extra labeled data is used for training in TASTR, therefore it is more scalable than those approaches that require extra Re-ID dataset or attribute labels. %In all methods, no spatio-temporal information is used in testing phase, since they may benefit from it on short-time dataset, but would risk to introduce error on long-time dataset.

\begin{figure}
	\begin{center}
		\includegraphics[width=1.0\linewidth]{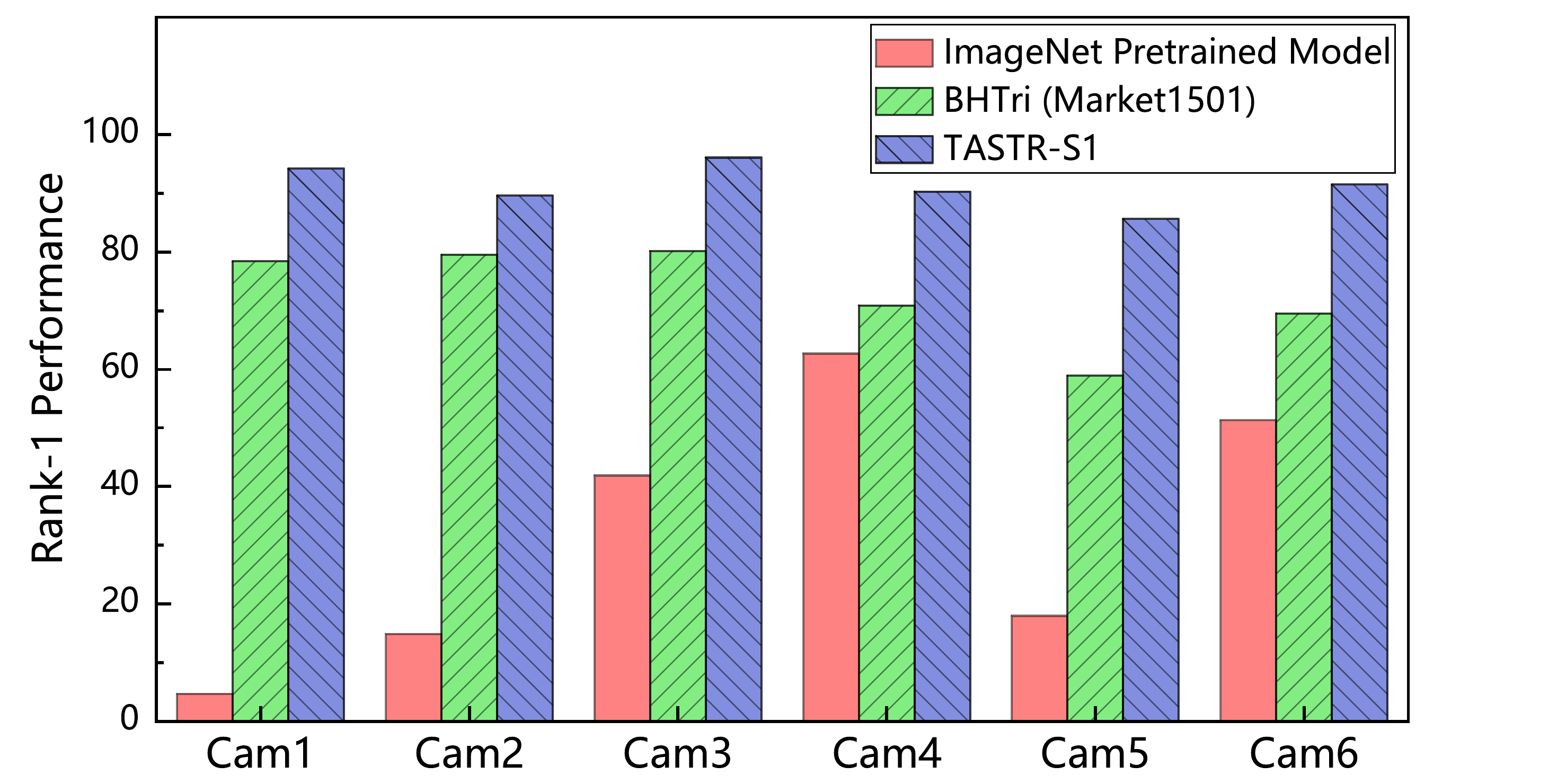}
	\end{center}
	\caption{Illustration of the within-view ID discrimination capability (rank-1 performance) of different models on different cameras on Campus-4K dataset.}
	\label{fig:s1}
\end{figure}

\begin{figure*}
	\begin{center}
		\subfigure[Campus-4K]{
			\includegraphics[width=0.48\linewidth]{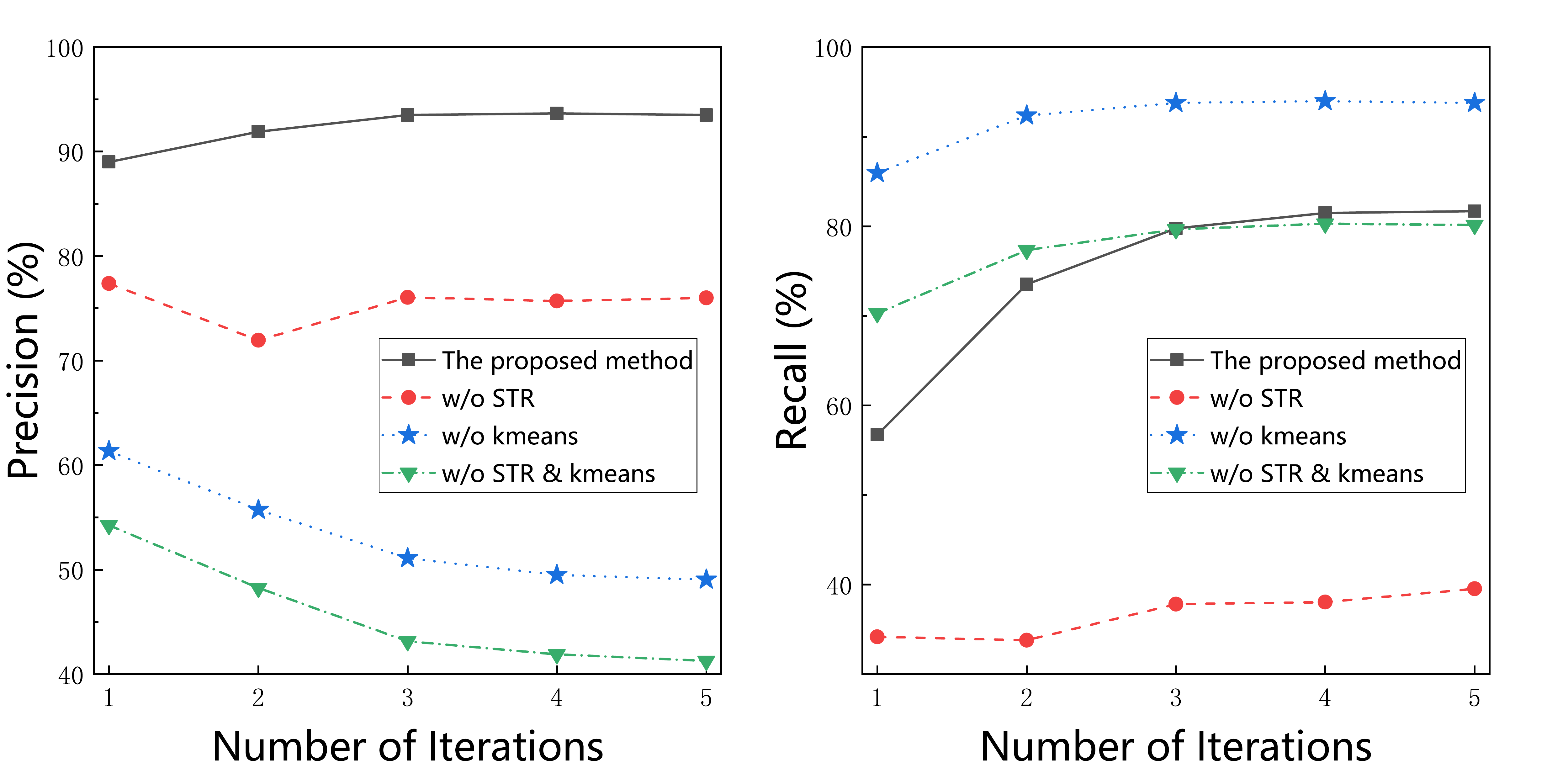}
		}
		\subfigure[DukeMTMC-reID]{
			\includegraphics[width=0.48\linewidth]{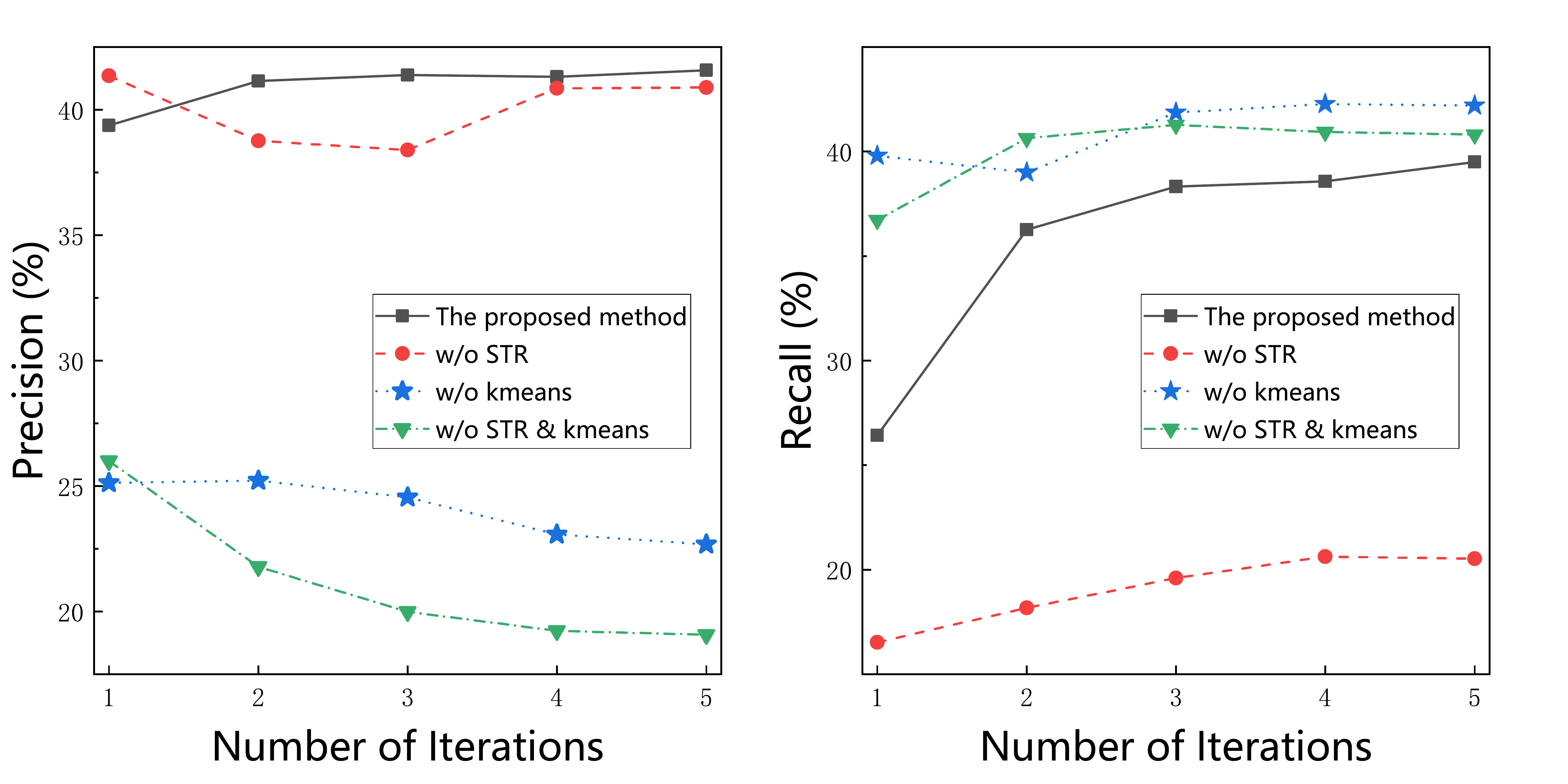}
		}
	\end{center}
	\caption{The precision and recall of the cross-camera associated tracklet pairs \emph{w.r.t.} different iterations of progressive optimization on (a) Campus-4K and (b) DukeMTMC-reID}
	\label{fig:error}
\end{figure*} 

%\newcommand{\tabincell}[2]{\begin{tabular}{@{}#1@{}}#2\end{tabular}}
%\begin{table}
%	%	\footnotesize
%	\begin{center}
%		%		\resizebox{0.48\textwidth}{!}{%
%		\begin{tabular}{c|c|c|c}
%			\hline
%			Models & \tabincell{c}{ImageNet \\ Pretrained}  & \tabincell{c}{BHTri \\ (Market1501)}  & \textbf{TASTR-S1} \\
%			\hline\hline
%			Cam1 & 4.7 & 78.4 & \textbf{94.2} \\
%			Cam2 & 14.9 & 79.5 & \textbf{89.6} \\
%			Cam3 & 41.9 & 80.1 & \textbf{96.1} \\
%			Cam4 & 62.7 & 70.9 & \textbf{90.3} \\
%			Cam5 & 18.0 & 58.9 & \textbf{85.6} \\
%			Cam6 & 51.3 & 69.5 & \textbf{91.5} \\
%			\hline
%		\end{tabular}%}%
%	\end{center}
%	\caption{Illustration of the within-view ID discrimination capability of different models on different cameras (rank-1).}
%	\label{tab:within}
%\end{table}

\subsection{Ablation Study}

We evaluate the effectiveness of different components as shown in Table.~\ref{tab:component}, which are categorized into three groups, \emph{i.e.}, 1) batch hard triplet loss based model trained on Market1501 (make prediction directly) and target dataset, 2) within-camera Re-ID training with or without the TCSCC, and 3) cross-camera Re-ID training by 1-reciprocal Nearest Neighbor based Cross-Camera Tracklets Association (NN-CCTA) with different refinements. 

\subsubsection{Effectiveness of the Training Data in Target Domain}\label{sec:s1}
From group 1 we can see that directly deploy the model pre-trained on another dataset to a new dataset will lead to poor performance. And the performance of TASTR-S1 in group 2, which only uses within-camera tracking data for training, is significantly superior to the pre-trained model on Market1501 by a large margin. On the one hand, it is due to the fact that the target domain can be significantly different from the source domain. On the other hand, limited scale of current Re-ID datasets restricts the generalization capability of the model. However, as mentioned at the beginning, annotating the target data from online surveillance videos of large-scale camera network is strenuous and impractical.
It reveals the importance of unsupervised methods that can take advantage of the data obtained in the target environment.

\subsubsection{Effectiveness of Within-Camera Re-ID Training}
Within-camera person tracklets can provide some ID discriminative information for Re-ID learning. However, there are two problems for a pure unsupervised method, \emph{i.e.}, within-camera ID duplication and cross-camera ID duplication. The results in group 2 of Table~\ref{tab:component} show that without TCSCC sampling method, \emph{i.e.}, for each triplet batch only the samples from the same camera will be selected and there should be a time gap larger than $T$ between them, Triplet based Re-ID model may achieve very poor performance, while hard mining may degrade it since different tracklets of the same person are more likely to be hard negatives and contribute to the loss, thus guide the Re-ID learning in a wrong way.

Besides, we conduct an additional experiment to evaluate the within-view ID discrimination capability of TASTR-S1 on Campus-4K dataset. For all tracklets in testing set, we randomly remove several frames from the middle third of them to simulate tracking fragmentation. So each tracklet is divided into two sub-tracklets, then we put one into query set and another into gallery set to verify whether the model can rematch them. The rank-1 performances of different models are shown in Fig.~\ref{fig:s1}, and TASTR-S1 is much stronger than other models pre-trained on ImageNet or Market1501. Although the actual situation in real-world application is more complicated, it is worth pointing out that all tracklet data for TASTR-S1 training come from single-camera multi-person tracking, and TASTR-S1 demonstrates great potential for promoting the performance of single-camera multi-person tracking in return. 

\subsubsection{Effectiveness of Cross-Camera Re-ID Training}
\begin{figure*}
	\begin{center}
		\subfigure[Campus-4K]{
			\includegraphics[width=1.0\linewidth]{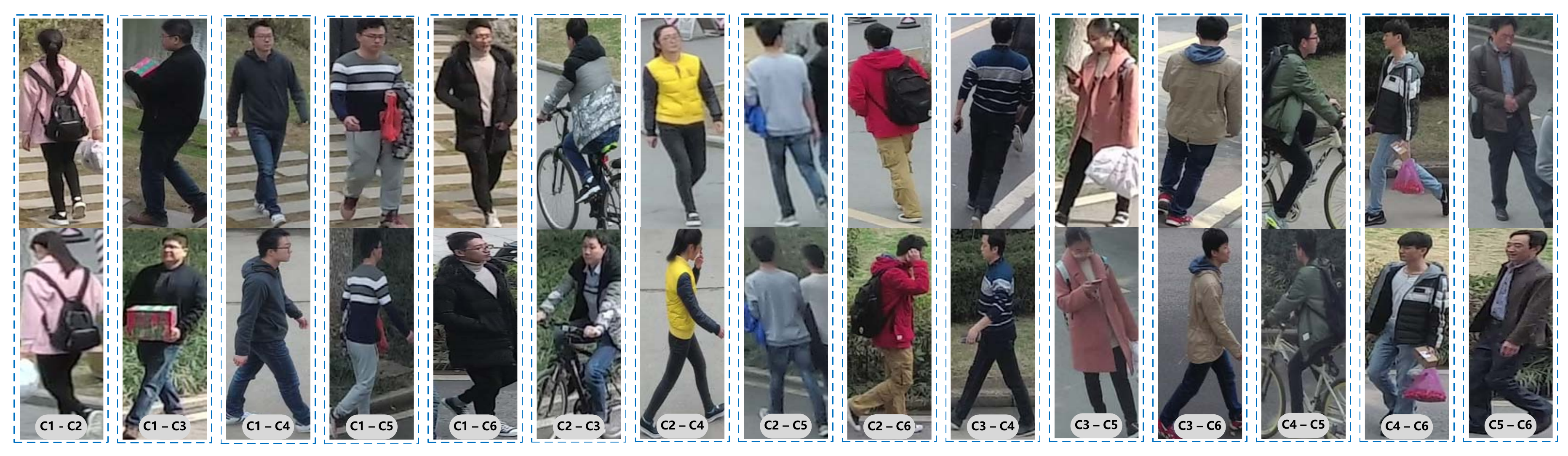}
		}
		\subfigure[DukeMTMC-reID]{
			\includegraphics[width=1.0\linewidth]{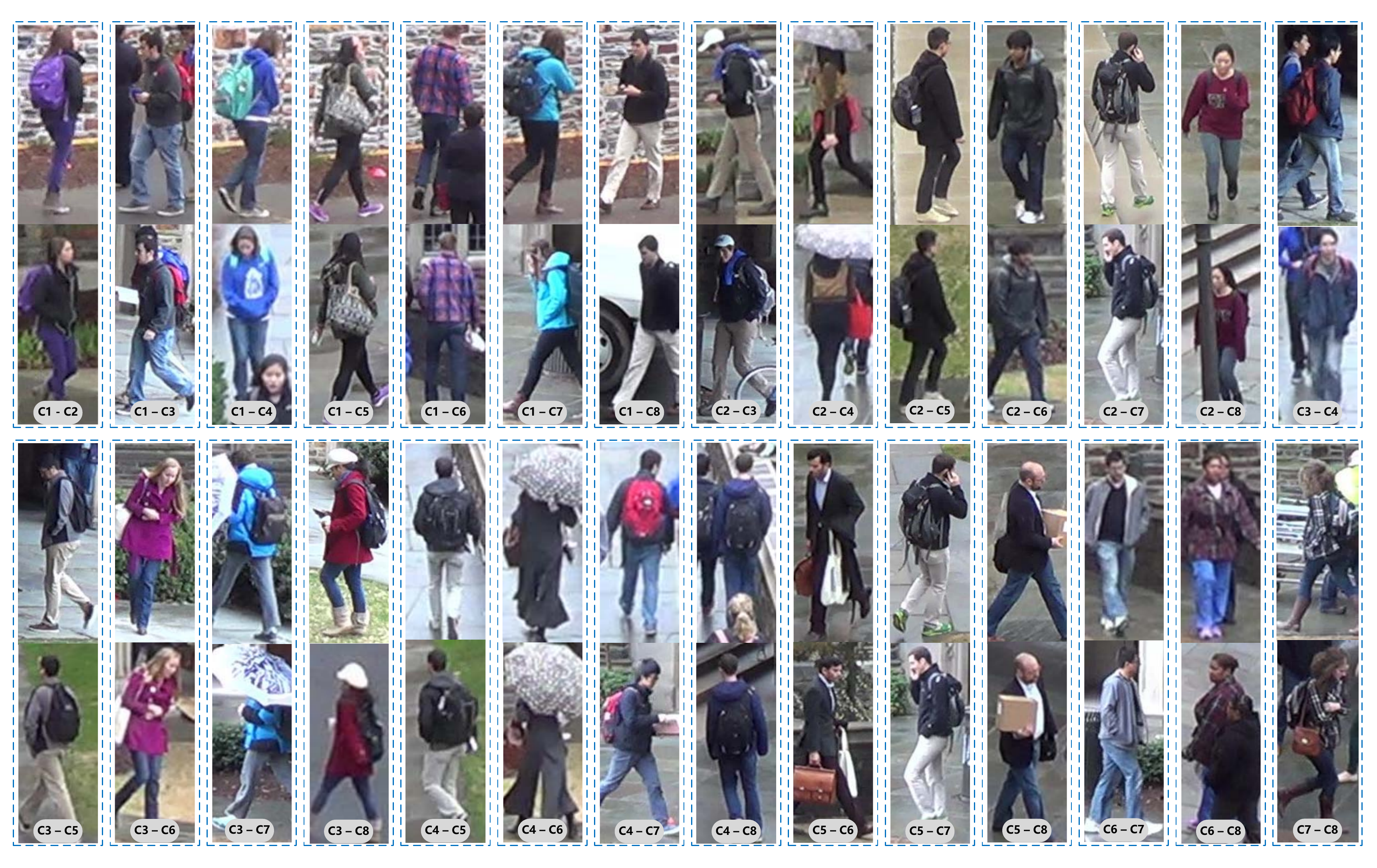}
		}
	\end{center}
	\caption{Example cross-camera matching tracklet pairs obtained by the proposed method on (a) Campus-4K, (b) DukeMTMC-reID. ``C1-C2'': Matching pairs between Camera1 and Camera2. }
	\label{fig:matches}
\end{figure*}

As shown in group 3 of Table~\ref{tab:component}. Based on TASTR-S1, we conduct 1-reciprocal Nearest Neighbor based Cross-Camera Tracklet Association (NN-CCTA) for cross-view tracklet matching. Even without Spatio-Temporal Regularization (STR) and $k$-means, NN-CCTA can get a big improvement over TASTR-S1 because of the importance of cross-view ID discriminative learning. However, when performing it in a progressive optimization way, the performance gain is limited. This is caused by the precision and quantity of cross-camera matching pairs, which makes the model unstable. If we make some refining such as STR or $k$-means, the performances may be lower than NN-CCTA. The main reason is that a large number of matching tracklets with potential information useful for more effective learning are temporarily abandoned. However, these operations have greater potential for progressive optimization and TASTR (with STR \& k-means) achieves the best performance. It indicates that further refining the cross-camera tracklet association results may be significant for progressive optimization as errors can be propagated from wrong matching pairs. Fig.~\ref{fig:matches} shows some examples of cross-camera matching tracklet pairs obtained by the final TASTR model on Campus-4K and DukeMTMC-reID.

\begin{table}
	%	\footnotesize
	\begin{center}
		%		\resizebox{0.48\textwidth}{!}{%
		\begin{tabular}{c|cccc}
			\hline
			Models & rank-1  & rank-5  & rank-10 & mAP \\
			\hline\hline
			TASTR-S1 (unsupervised) & 53.0 & 74.7 & 83.2 & 53.9 \\
			TASTR-S1 (weakly supervised) & 58.6 & 78.6 & 84.4 & 57.8 \\
			\hline
			TASTR (unsupervised) & 76.4 & 91.6 & 94.6 & 78.3 \\
			TASTR (weakly supervised) & \textbf{82.7} & \textbf{94.6} & \textbf{96.8} & \textbf{84.4} \\
			\hline
		\end{tabular}%}%
	\end{center}
	\caption{Evaluation of weakly supervised learning on Campus-4K}
	\label{tab:weakly}
\end{table}

\begin{figure*}
	\begin{center}
		\subfigure[Campus-4K]{
			\includegraphics[width=0.45\linewidth]{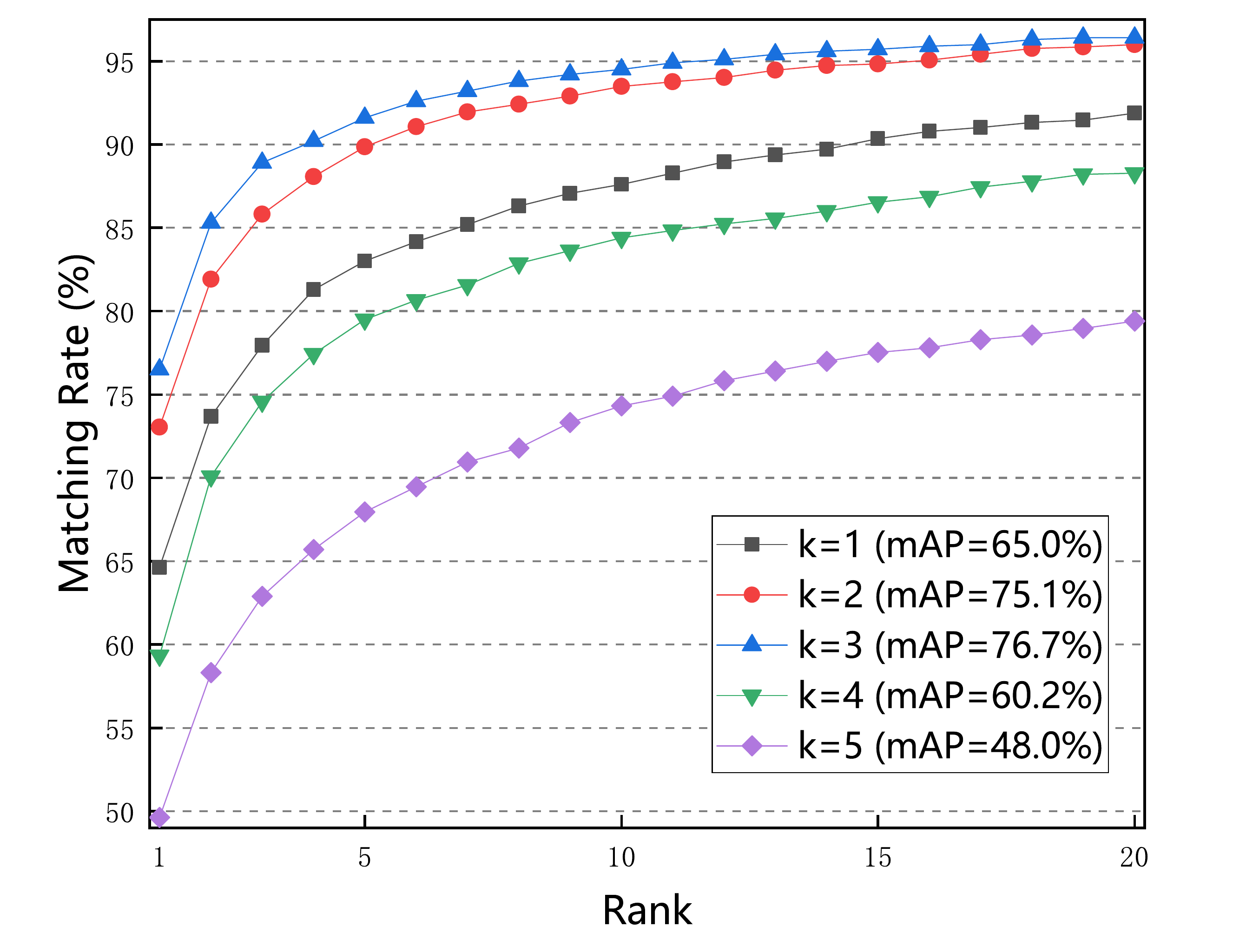}
		}
		\subfigure[DukeMTMC-reID]{
			\includegraphics[width=0.45\linewidth]{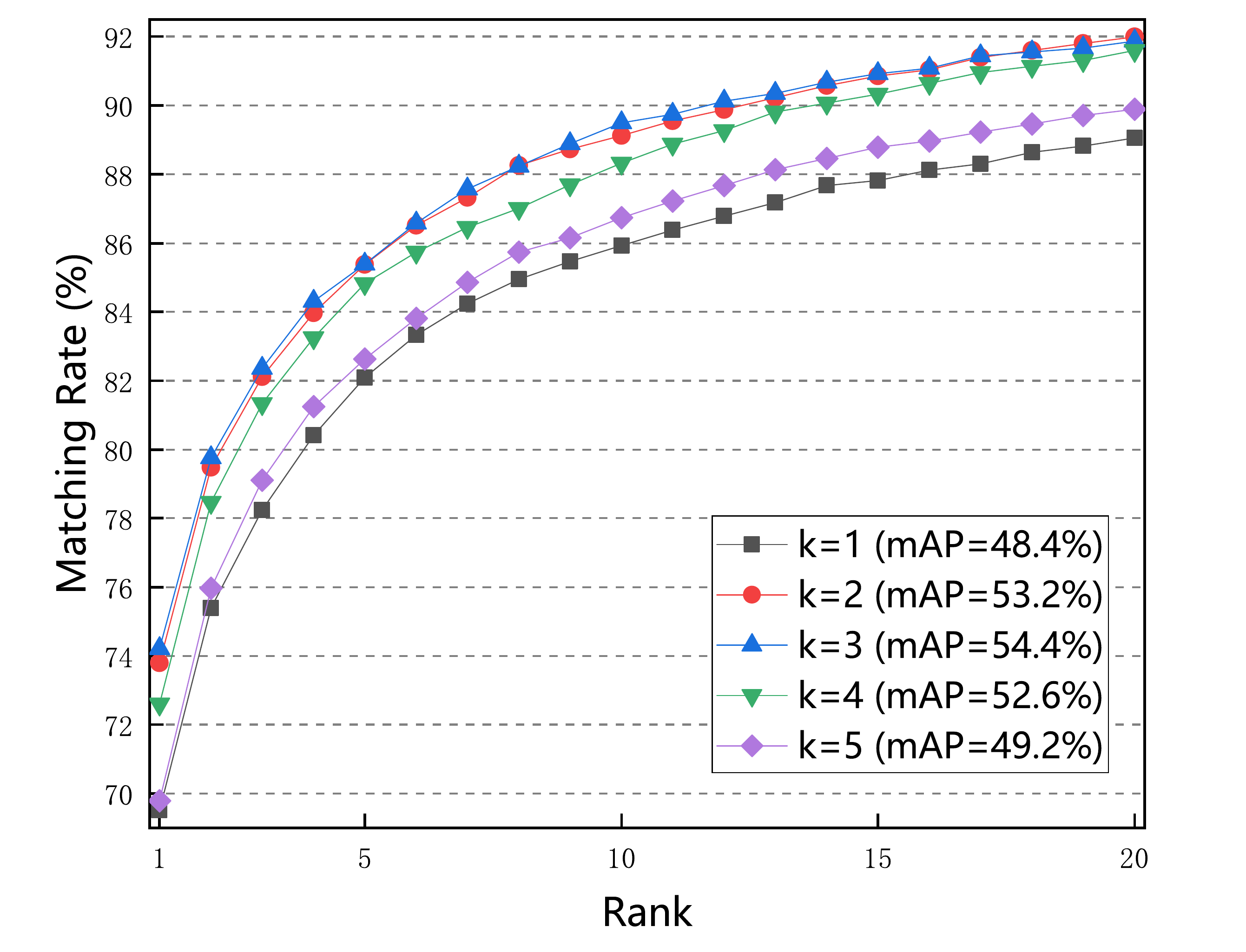}
		}
	\end{center}
	\caption{CMC curves and mAP corresponding to different $k$ in $k$-means on (a) Campus-4K and (b) DukeMTMC-reID}
	\label{fig:k}
\end{figure*}

\begin{figure}
	\begin{center}
		\subfigure[Campus-4K]{
			\includegraphics[width=0.46\linewidth]{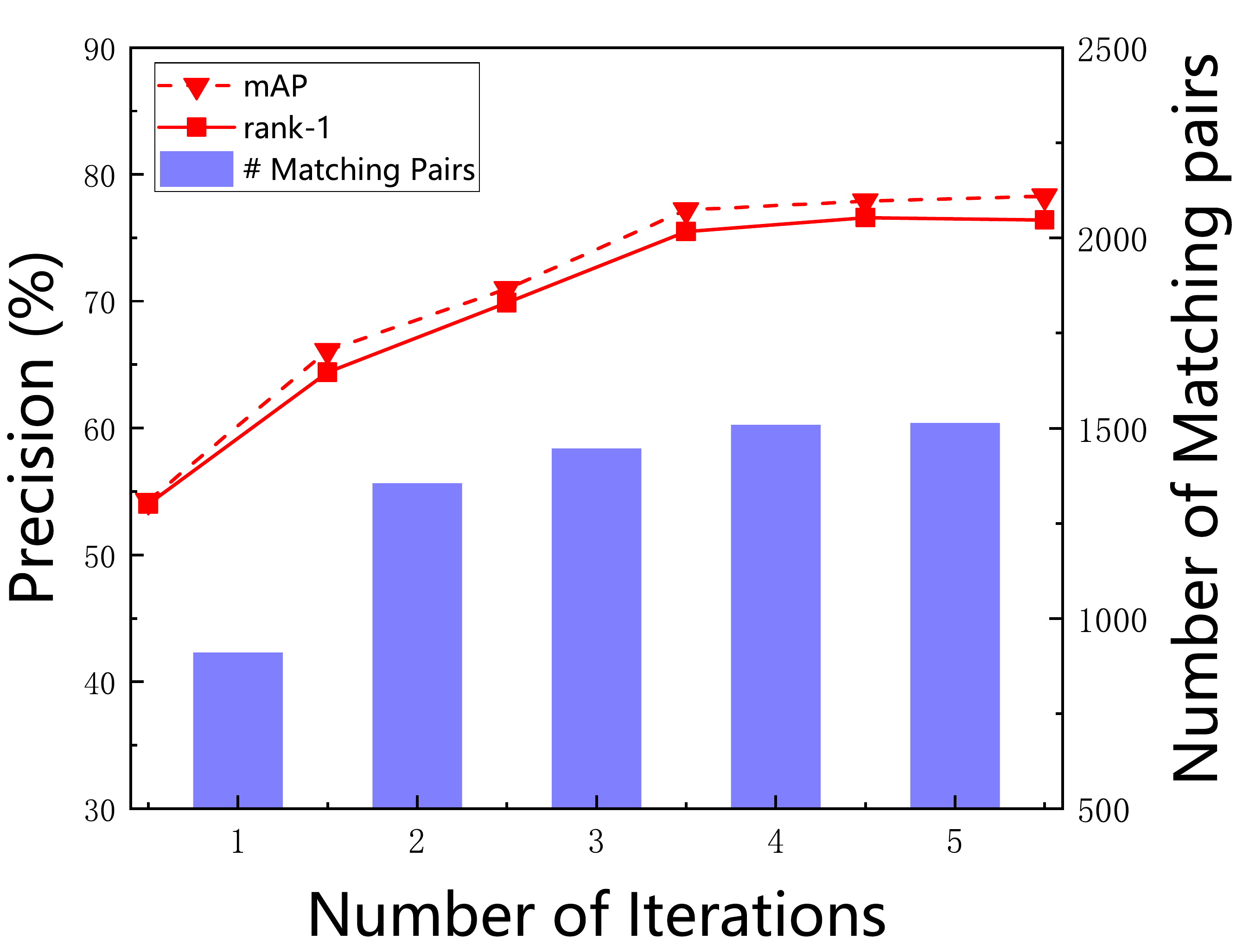}
		}
		\subfigure[DukeMTMC-reID]{
			\includegraphics[width=0.46\linewidth]{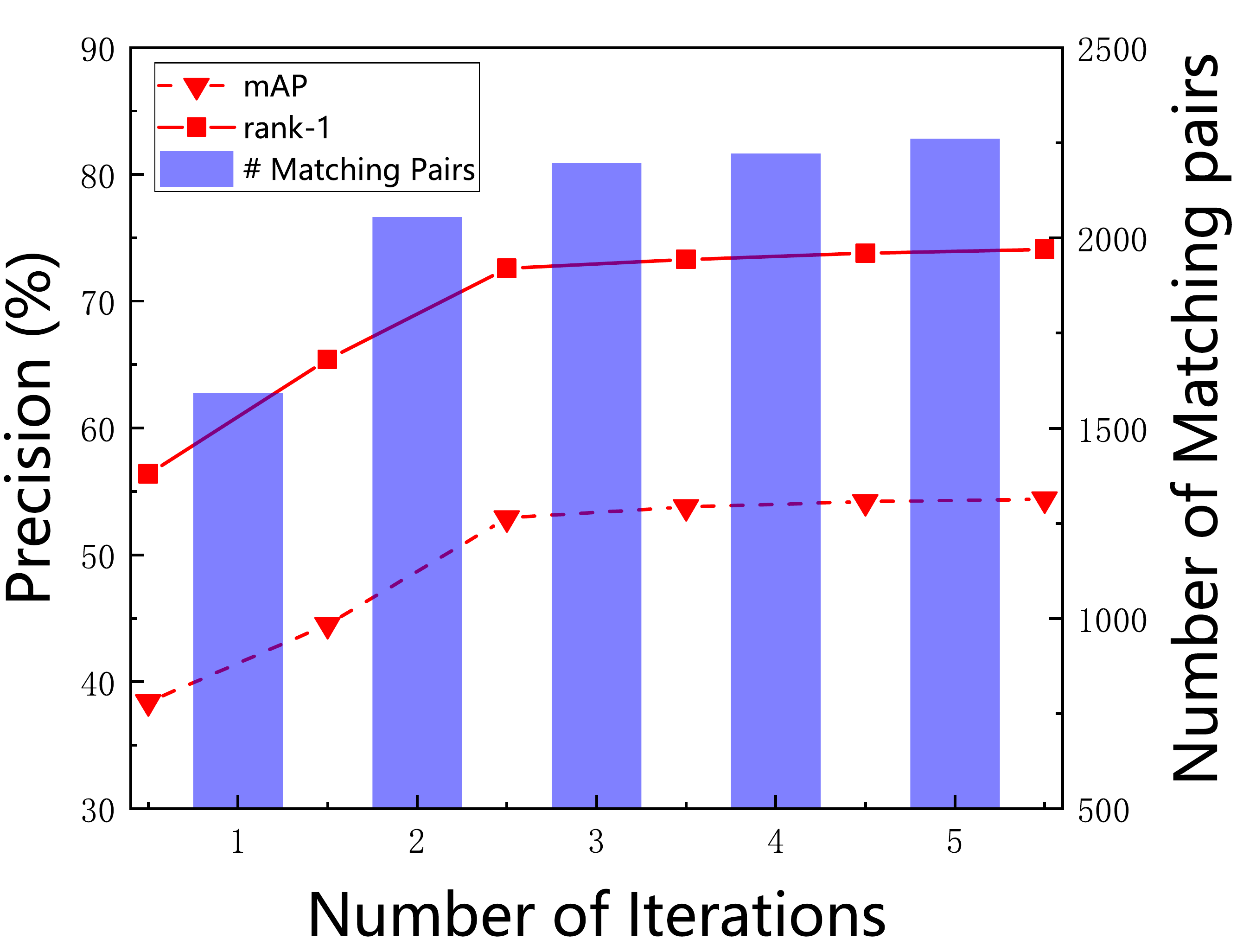}
		}
	\end{center}
	\caption{The performance and the number of associated tracklet pairs \emph{w.r.t.} different iterations (0 means TASTR-S1) of progressive optimization on (a) Campus-4K and (b) DukeMTMC-reID}
	\label{fig:iter}
\end{figure}

\subsection{Weakly Supervised Learning}

In person Re-ID training data annotation, the process can be divided into two steps: 1) per-camera person annotation, 2) cross-camera person matching. The first step is to track pedestrians manually, which is quite easy for human. The second step requires exhaustive manual search of cross-camera person matching, which is the most costly procedure as it is hard to know when and where a specific person will appear given complex camera network topology and unconstrained people's moving in the public spaces. Thus, per-camera ID labeling is much cheaper, and such labeled data are much weaker due to the lack of cross-camera ID pairs information. This setting in person Re-ID can be called weakly supervised learning.

The proposed TASTR can be easily applied in the weakly supervised setting. In this setting, per-camera person ID information is given, so there is almost no within-camera ID duplication problem due to no trajectory fragmentation. Furthermore, it allows to test how much performance gain the Re-ID model can get from perfect within-camera multi-person detection and tracking. Table~\ref{tab:weakly} shows the results of TASTR-S1 and TASTR under weakly supervised setting on Campus-4K. This demonstrates the suitability of our method for different labeling settings in real-world scenarios.

\subsection{Error Analysis of Tracklet Association}

%\begin{figure}
%	\centering
%	\includegraphics[width=1.0\linewidth]{acc_recall.pdf}
%	\caption{The precision and recall of associated tracklet pairs on DukeMTMC-reID}
%	\label{fig:iter}
%\end{figure}

The precision and recall of the associated tracklet pairs over iterations are shown in Fig.~\ref{fig:error}. Since no label information is available on Campus-4K when conducting pure unsupervised learning from automatically generated person tracklet data, the above results are obtained under weakly supervised setting. Without $k$-means and STR, the model would get relatively high recall but the lowest precision. STR (\emph{i.e.} w/o $k$-means) can improve both the precision and recall rate of NN-CCTA. $k$-means removes a large number of matching tracklet pairs with large distances, which results in higher precision but lowest recall. The proposed method TASTR (NN-CCTA with STR and $k$-means) achieves the highest precision as well as high recall rate, and the recall grows steadily while the precision remains at a high level in the process of progressive optimization, which leads to the best Re-ID performance. This observation demonstrates the effectiveness of our method, which can acquire more training data while achieving higher precision.

\subsection{Parameter Sensitivity Study}

In our TASTR, there are two important parameters, \emph{i.e.}, $k$ in $k$-means clustering and $n$ which denotes the iteration number of progressive optimization. In the following, we study the sensitivity of our approach to the setting of the two parameters.  

As shown in Fig.~\ref{fig:k}, when $k=1$, which means all matching tracklet pair candidates are used for progressive training, the performance of Re-ID model is poor as it introduces more incorrectly matching pairs. With $k$ increases, it becomes more and more strict for cross-camera tracklet association, thus leads to fewer and fewer training data for cross-view discriminative learning. The best performance is obtained when $k=3$, which indicates a balance between the accuracy of cross-camera tracklet matching and the sufficiency of cross-view training data.

To evaluate the effectiveness of progressive optimization, we train our model with the best setting and observe its performance and the number of associated tracklet pairs in different iterations. The mAP and rank-1 performance before the first iteration means the performance of TASTR-S1 model. As shown in Fig.~\ref{fig:iter}, both the performance and the number of associated tracklet pairs gain considerable improvement in the first three iterations of cross-view optimization, and then keeps relatively stable. In order to get the best results, we set $n=5$ in our approach.

\section{Conclusion}
In this work, we are dedicated to unsupervised person Re-ID and contribute a new high-quality Campus-4K dataset with full frames and full spatio-temporal information. We propose a new progressive learning method by Tracklet Association with Spatio-Temporal Regularization (TASTR). Instead of assuming perfect person tracking in each camera view, we automatically explore the identity discriminative information from imperfect tracklets with spatio-temporal context. We initially learn a Re-ID model with triplets constructed with spatio-temporal constraint. Then, we associate the tracklets across each camera pair which is further used to fine-tune and update the initial Re-ID model. The above processes are iterated to progressively improve the discriminative capability of the Re-ID model. We make evaluations on two datasets and extensive experiments demonstrate the effectiveness of our approach.

% or


%\appendices

%\section{}

%\section*{Acknowledgment}

\ifCLASSOPTIONcaptionsoff
  \newpage
\fi

%\emph{i.e.}EEtriggeratref{8}
%\emph{i.e.}EEtriggercmd{\enlargethispage{-5in}}


%\bibliographystyle{IEEEtran}

\bibliographystyle{IEEEtran}
\bibliography{egbib}

\begin{thebibliography}{10}
\providecommand{\url}[1]{#1}
\csname url@samestyle\endcsname
\providecommand{\newblock}{\relax}
\providecommand{\bibinfo}[2]{#2}
\providecommand{\BIBentrySTDinterwordspacing}{\spaceskip=0pt\relax}
\providecommand{\BIBentryALTinterwordstretchfactor}{4}
\providecommand{\BIBentryALTinterwordspacing}{\spaceskip=\fontdimen2\font plus
\BIBentryALTinterwordstretchfactor\fontdimen3\font minus
  \fontdimen4\font\relax}
\providecommand{\BIBforeignlanguage}[2]{{%
\expandafter\ifx\csname l@#1\endcsname\relax
\typeout{** WARNING: IEEEtran.bst: No hyphenation pattern has been}%
\typeout{** loaded for the language `#1'. Using the pattern for}%
\typeout{** the default language instead.}%
\else
\language=\csname l@#1\endcsname
\fi
#2}}
\providecommand{\BIBdecl}{\relax}
\BIBdecl

\bibitem{li2014deepreid}
W.~Li, R.~Zhao, T.~Xiao, and X.~Wang, ``Deepreid: Deep filter pairing neural
  network for person re-identification,'' in \emph{Proceedings of the IEEE
  Conference on Computer Vision and Pattern Recognition (CVPR)}, 2014, pp.
  152--159.

\bibitem{wang2015zero}
Z.~Wang, R.~Hu, C.~Liang, Y.~Yu, J.~Jiang, M.~Ye, J.~Chen, and Q.~Leng,
  ``Zero-shot person re-identification via cross-view consistency,'' \emph{IEEE
  Transactions on Multimedia (TMM)}, vol.~18, no.~2, pp. 260--272, 2015.

\bibitem{ahmed2015improved}
E.~Ahmed, M.~Jones, and T.~K. Marks, ``An improved deep learning architecture
  for person re-identification,'' in \emph{Proceedings of the IEEE Conference
  on Computer Vision and Pattern Recognition (CVPR)}, 2015, pp. 3908--3916.

\bibitem{ye2016person}
M.~Ye, C.~Liang, Y.~Yu, Z.~Wang, Q.~Leng, C.~Xiao, J.~Chen, and R.~Hu, ``Person
  reidentification via ranking aggregation of similarity pulling and
  dissimilarity pushing,'' \emph{IEEE Transactions on Multimedia (TMM)},
  vol.~18, no.~12, pp. 2553--2566, 2016.

\bibitem{chen2017beyond}
W.~Chen, X.~Chen, J.~Zhang, and K.~Huang, ``Beyond triplet loss: a deep
  quadruplet network for person re-identification,'' in \emph{Proceedings of
  the IEEE Conference on Computer Vision and Pattern Recognition (CVPR)}, 2017,
  pp. 403--412.

\bibitem{Zhao_2017_ICCV}
L.~Zhao, X.~Li, Y.~Zhuang, and J.~Wang, ``Deeply-learned part-aligned
  representations for person re-identification,'' in \emph{Proceedings of the
  IEEE International Conference on Computer Vision (ICCV)}, 2017, pp.
  3219--3228.

\bibitem{hermans2017defense}
A.~Hermans, L.~Beyer, and B.~Leibe, ``In defense of the triplet loss for person
  re-identification,'' \emph{arXiv preprint arXiv:1703.07737}, 2017.

\bibitem{liu2017end}
H.~Liu, J.~Feng, M.~Qi, J.~Jiang, and S.~Yan, ``End-to-end comparative
  attention networks for person re-identification,'' \emph{IEEE Transactions on
  Image Processing (TIP)}, vol.~26, no.~7, pp. 3492--3506, 2017.

\bibitem{zhou2017large}
S.~Zhou, J.~Wang, R.~Shi, Q.~Hou, Y.~Gong, and N.~Zheng, ``Large margin
  learning in set-to-set similarity comparison for person reidentification,''
  \emph{IEEE Transactions on Multimedia (TMM)}, vol.~20, no.~3, pp. 593--604,
  2017.

\bibitem{li2018harmonious}
W.~Li, X.~Zhu, and S.~Gong, ``Harmonious attention network for person
  re-identification,'' in \emph{Proceedings of the IEEE Conference on Computer
  Vision and Pattern Recognition (CVPR)}, 2018, pp. 2285--2294.

\bibitem{wei2018person}
L.~Wei, S.~Zhang, W.~Gao, and Q.~Tian, ``Person transfer gan to bridge domain
  gap for person re-identification,'' in \emph{Proceedings of the IEEE
  Conference on Computer Vision and Pattern Recognition (CVPR)}, 2018, pp.
  79--88.

\bibitem{song2018mask}
C.~Song, Y.~Huang, W.~Ouyang, and L.~Wang, ``Mask-guided contrastive attention
  model for person re-identification,'' in \emph{Proceedings of the IEEE
  Conference on Computer Vision and Pattern Recognition (CVPR)}, 2018, pp.
  1179--1188.

\bibitem{wang2019incremental}
Z.~Wang, J.~Jiang, Y.~Yu, and S.~Satoh, ``Incremental re-identification by
  cross-direction and cross-ranking adaption,'' \emph{IEEE Transactions on
  Multimedia (TMM)}, 2019.

\bibitem{zheng2019pose}
L.~Zheng, Y.~Huang, H.~Lu, and Y.~Yang, ``Pose invariant embedding for deep
  person re-identification,'' \emph{IEEE Transactions on Image Processing
  (TIP)}, 2019.

\bibitem{fan2018unsupervised}
H.~Fan, L.~Zheng, C.~Yan, and Y.~Yang, ``Unsupervised person re-identification:
  Clustering and fine-tuning,'' \emph{ACM Transactions on Multimedia Computing,
  Communications, and Applications (TOMM)}, vol.~14, no.~4, p.~83, 2018.

\bibitem{lv2018unsupervised}
J.~Lv, W.~Chen, Q.~Li, and C.~Yang, ``Unsupervised cross-dataset person
  re-identification by transfer learning of spatial-temporal patterns,'' in
  \emph{Proceedings of the IEEE Conference on Computer Vision and Pattern
  Recognition (CVPR)}, 2018, pp. 7948--7956.

\bibitem{kodirov2015dictionary}
E.~Kodirov, T.~Xiang, and S.~Gong, ``Dictionary learning with iterative
  laplacian regularisation for unsupervised person re-identification,'' in
  \emph{Proceedings of the British Machine Vision Conference (BMVC)}, vol.~3,
  2015, p.~8.

\bibitem{wang2016towards}
H.~Wang, X.~Zhu, T.~Xiang, and S.~Gong, ``Towards unsupervised open-set person
  re-identification,'' in \emph{Proceedings of the IEEE International
  Conference on Image Processing (ICIP)}, 2016, pp. 769--773.

\bibitem{kodirov2016person}
E.~Kodirov, T.~Xiang, Z.~Fu, and S.~Gong, ``Person re-identification by
  unsupervised $l_1$ graph learning,'' in \emph{Proceedings of the European
  Conference on Computer Vision (ECCV)}, 2016, pp. 178--195.

\bibitem{zhao2017person}
R.~Zhao, W.~Oyang, and X.~Wang, ``Person re-identification by saliency
  learning,'' \emph{IEEE Transactions on Pattern Analysis and Machine
  Intelligence (TPAMI)}, vol.~39, no.~2, pp. 356--370, 2017.

\bibitem{Liu_2017_ICCV}
Z.~Liu, D.~Wang, and H.~Lu, ``Stepwise metric promotion for unsupervised video
  person re-identification,'' in \emph{Proceedings of the IEEE International
  Conference on Computer Vision (ICCV)}, 2017, pp. 2429--2438.

\bibitem{Ye_2017_ICCV}
M.~Ye, A.~J. Ma, L.~Zheng, J.~Li, and P.~C. Yuen, ``Dynamic label graph
  matching for unsupervised video re-identification,'' in \emph{Proceedings of
  the IEEE International Conference on Computer Vision (ICCV)}, 2017, pp.
  5142--5150.

\bibitem{ma2017person}
X.~Ma, X.~Zhu, S.~Gong, X.~Xie, J.~Hu, K.-M. Lam, and Y.~Zhong, ``Person
  re-identification by unsupervised video matching,'' \emph{Pattern Recognition
  (PR)}, vol.~65, pp. 197--210, 2017.

\bibitem{huang2016camera}
W.~Huang, R.~Hu, C.~Liang, Y.~Yu, Z.~Wang, X.~Zhong, and C.~Zhang, ``Camera
  network based person re-identification by leveraging spatial-temporal
  constraint and multiple cameras relations,'' in \emph{International
  Conference on Multimedia Modeling (MMM)}, 2016, pp. 174--186.

\bibitem{martinel2017person}
N.~Martinel, G.~L. Foresti, and C.~Micheloni, ``Person reidentification in a
  distributed camera network framework,'' \emph{IEEE transactions on
  cybernetics}, vol.~47, no.~11, pp. 3530--3541, 2017.

\bibitem{cho2019joint}
Y.-J. Cho, S.-A. Kim, J.-H. Park, K.~Lee, and K.-J. Yoon, ``Joint person
  re-identification and camera network topology inference in multiple
  cameras,'' \emph{Computer Vision and Image Understanding}, vol. 180, pp.
  34--46, 2019.

\bibitem{wang2018spatial}
G.~Wang, J.~Lai, P.~Huang, and X.~Xie, ``Spatial-temporal person
  re-identification,'' \emph{AAAI Conference on Artificial Intelligence
  (AAAI)}, 2019.

\bibitem{wu2018exploit}
Y.~Wu, Y.~Lin, X.~Dong, Y.~Yan, W.~Ouyang, and Y.~Yang, ``Exploit the unknown
  gradually: One-shot video-based person re-identification by stepwise
  learning,'' in \emph{Proceedings of the IEEE Conference on Computer Vision
  and Pattern Recognition (CVPR)}, 2018, pp. 5177--5186.

\bibitem{farenzena2010person}
M.~Farenzena, L.~Bazzani, A.~Perina, V.~Murino, and M.~Cristani, ``Person
  re-identification by symmetry-driven accumulation of local features,'' in
  \emph{Proceedings of the IEEE Conference on Computer Vision and Pattern
  Recognition (CVPR)}, 2010, pp. 2360--2367.

\bibitem{ma2012bicov}
B.~Ma, Y.~Su, and F.~Jurie, ``Bicov: a novel image representation for person
  re-identification and face verification,'' in \emph{Proceedings of the
  British Machine Vision Conference (BMVC)}, 2012.

\bibitem{ma2012local}
B.~Ma, Y.~Su, and F.~Jurie, ``Local descriptors encoded by fisher vectors for person
  re-identification,'' in \emph{Proceedings of the European Conference on
  Computer Vision (ECCV)}, 2012, pp. 413--422.

\bibitem{liu2014semi}
X.~Liu, M.~Song, D.~Tao, X.~Zhou, C.~Chen, and J.~Bu, ``Semi-supervised coupled
  dictionary learning for person re-identification,'' in \emph{Proceedings of
  the IEEE Conference on Computer Vision and Pattern Recognition (CVPR)}, 2014,
  pp. 3550--3557.

\bibitem{zhao2013unsupervised}
R.~Zhao, W.~Ouyang, and X.~Wang, ``Unsupervised salience learning for person
  re-identification,'' in \emph{Proceedings of the IEEE Conference on Computer
  Vision and Pattern Recognition (CVPR)}, 2013, pp. 3586--3593.

\bibitem{peng2016unsupervised}
P.~Peng, T.~Xiang, Y.~Wang, M.~Pontil, S.~Gong, T.~Huang, and Y.~Tian,
  ``Unsupervised cross-dataset transfer learning for person
  re-identification,'' in \emph{Proceedings of the IEEE Conference on Computer
  Vision and Pattern Recognition (CVPR)}, 2016, pp. 1306--1315.

\bibitem{zhu2017unpaired}
J.-Y. Zhu, T.~Park, P.~Isola, and A.~A. Efros, ``Unpaired image-to-image
  translation using cycle-consistent adversarial networks,'' in
  \emph{Proceedings of the IEEE International Conference on Computer Vision
  (ICCV)}, 2017, pp. 2223--2232.

\bibitem{su2016deep}
C.~Su, S.~Zhang, J.~Xing, W.~Gao, and Q.~Tian, ``Deep attributes driven
  multi-camera person re-identification,'' in \emph{Proceedings of the European
  Conference on Computer Vision (ECCV)}, 2016, pp. 475--491.

\bibitem{wang2018transferable}
J.~Wang, X.~Zhu, S.~Gong, and W.~Li, ``Transferable joint attribute-identity
  deep learning for unsupervised person re-identification,'' in
  \emph{Proceedings of the IEEE Conference on Computer Vision and Pattern
  Recognition (CVPR)}, 2018, pp. 2275--2284.

\bibitem{deng2018image}
W.~Deng, L.~Zheng, Q.~Ye, G.~Kang, Y.~Yang, and J.~Jiao, ``Image-image domain
  adaptation with preserved self-similarity and domain-dissimilarity for person
  re-identification,'' in \emph{Proceedings of the IEEE Conference on Computer
  Vision and Pattern Recognition (CVPR)}, 2018, pp. 994--1003.

\bibitem{li2018unsupervised}
M.~Li, X.~Zhu, and S.~Gong, ``Unsupervised person re-identification by deep
  learning tracklet association,'' in \emph{Proceedings of the European
  Conference on Computer Vision (ECCV)}, 2018, pp. 737--753.

\bibitem{gray2008viewpoint}
D.~Gray and H.~Tao, ``Viewpoint invariant pedestrian recognition with an
  ensemble of localized features,'' in \emph{Proceedings of the European
  Conference on Computer Vision (ECCV)}, 2008, pp. 262--275.

\bibitem{hirzer2011person}
M.~Hirzer, C.~Beleznai, P.~M. Roth, and H.~Bischof, ``Person re-identification
  by descriptive and discriminative classification,'' in \emph{Scandinavian
  conference on Image analysis}, 2011, pp. 91--102.

\bibitem{zheng2015scalable}
L.~Zheng, L.~Shen, L.~Tian, S.~Wang, J.~Wang, and Q.~Tian, ``Scalable person
  re-identification: A benchmark,'' in \emph{Proceedings of the IEEE
  International Conference on Computer Vision (ICCV)}, 2015, pp. 1116--1124.

\bibitem{wang2016person}
T.~Wang, S.~Gong, X.~Zhu, and S.~Wang, ``Person re-identification by
  discriminative selection in video ranking,'' \emph{IEEE Transactions on
  Pattern Analysis and Machine Intelligence (TPAMI)}, vol.~38, no.~12, pp.
  2501--2514, 2016.

\bibitem{zheng2017unlabeled}
Z.~Zheng, L.~Zheng, and Y.~Yang, ``Unlabeled samples generated by gan improve
  the person re-identification baseline in vitro,'' in \emph{Proceedings of the
  IEEE International Conference on Computer Vision (ICCV)}, 2017, pp.
  3754--3762.

\bibitem{lan2018person}
X.~Lan, X.~Zhu, and S.~Gong, ``Person search by multi-scale matching,'' in
  \emph{Proceedings of the European Conference on Computer Vision (ECCV)},
  2018, pp. 536--552.

\bibitem{zheng2016person}
L.~Zheng, Y.~Yang, and A.~G. Hauptmann, ``Person re-identification: Past,
  present and future,'' \emph{arXiv preprint arXiv:1610.02984}, 2016.

\bibitem{fang2017rmpe}
H.-S. Fang, S.~Xie, Y.-W. Tai, and C.~Lu, ``Rmpe: Regional multi-person pose
  estimation,'' in \emph{Proceedings of the IEEE International Conference on
  Computer Vision (ICCV)}, 2017, pp. 2334--2343.

\bibitem{redmon2016you}
J.~Redmon, S.~Divvala, R.~Girshick, and A.~Farhadi, ``You only look once:
  Unified, real-time object detection,'' in \emph{Proceedings of the IEEE
  Conference on Computer Vision and Pattern Recognition (CVPR)}, 2016, pp.
  779--788.

\bibitem{xiu2018pose}
Y.~Xiu, J.~Li, H.~Wang, Y.~Fang, and C.~Lu, ``Pose flow: Efficient online pose
  tracking,'' \emph{Proceedings of the British Machine Vision Conference
  (BMVC)}, 2018.

\bibitem{ren2015faster}
S.~Ren, K.~He, R.~Girshick, and J.~Sun, ``Faster r-cnn: Towards real-time
  object detection with region proposal networks,'' in \emph{Advances in neural
  information processing systems}, 2015, pp. 91--99.

\bibitem{zhong2017re}
Z.~Zhong, L.~Zheng, D.~Cao, and S.~Li, ``Re-ranking person re-identification
  with k-reciprocal encoding,'' in \emph{Proceedings of the IEEE Conference on
  Computer Vision and Pattern Recognition (CVPR)}, 2017, pp. 1318--1327.

\bibitem{loy2009multi}
C.~C. Loy, T.~Xiang, and S.~Gong, ``Multi-camera activity correlation
  analysis,'' in \emph{Proceedings of the IEEE Conference on Computer Vision
  and Pattern Recognition (CVPR)}, 2009, pp. 1988--1995.

\bibitem{ristani2016performance}
E.~Ristani, F.~Solera, R.~Zou, R.~Cucchiara, and C.~Tomasi, ``Performance
  measures and a data set for multi-target, multi-camera tracking,'' in
  \emph{Proceedings of the European Conference on Computer Vision (ECCV)},
  2016, pp. 17--35.

\bibitem{he2016deep}
K.~He, X.~Zhang, S.~Ren, and J.~Sun, ``Deep residual learning for image
  recognition,'' in \emph{Proceedings of the IEEE Conference on Computer Vision
  and Pattern Recognition (CVPR)}, 2016, pp. 770--778.

\bibitem{kingma2014adam}
D.~P. Kingma and J.~Ba, ``Adam: A method for stochastic optimization,''
  \emph{arXiv preprint arXiv:1412.6980}, 2014.

\bibitem{liao2015person}
S.~Liao, Y.~Hu, X.~Zhu, and S.~Z. Li, ``Person re-identification by local
  maximal occurrence representation and metric learning,'' in \emph{Proceedings
  of the IEEE Conference on Computer Vision and Pattern Recognition (CVPR)},
  2015, pp. 2197--2206.

\bibitem{zhong2018generalizing}
Z.~Zhong, L.~Zheng, S.~Li, and Y.~Yang, ``Generalizing a person retrieval model
  hetero- and homogeneously,'' in \emph{Proceedings of the European Conference
  on Computer Vision (ECCV)}, 2018, pp. 172--188.

\bibitem{li2019unsupervised}
M.~Li, X.~Zhu, and S.~Gong, ``Unsupervised tracklet person re-identification,''
  \emph{IEEE Transactions on Pattern Analysis and Machine Intelligence
  (TPAMI)}, 2019.

\bibitem{yu2019unsupervised}
H.-X. Yu, W.-S. Zheng, A.~Wu, X.~Guo, S.~Gong, and J.-H. Lai, ``Unsupervised
  person re-identification by soft multilabel learning,'' in \emph{Proceedings
  of the IEEE Conference on Computer Vision and Pattern Recognition (CVPR)},
  2019, pp. 2148--2157.

\end{thebibliography}



% \end{thebibliography}


%\end{IEEEbiography}
%% insert where needed to balance the two columns on the last page with
%%\newpage


%\vfill


\end{document}